\documentclass[lettersize,journal]{IEEEtran}
\usepackage{enumitem,calc}
\usepackage{graphicx}
\usepackage{subfigure}
\usepackage{amsmath}
\usepackage{array}
\usepackage{subfig}
\usepackage{tabu}
\usepackage{amssymb}
\usepackage{multirow}
\usepackage{booktabs}
\usepackage{cite}
\usepackage{xcolor}
\usepackage{soul}
\soulregister\cite7
\soulregister\mathbf7

\usepackage{rotating}

\ifCLASSINFOpdf
\else
\fi
\begin{document}

\title{Quality-aware Selective Fusion Network for V-D-T Salient Object Detection}

\author{Liuxin Bao,
        Xiaofei Zhou*,
        Xiankai Lu,
        Yaoqi Sun,
        Haibing Yin,
        Zhenghui Hu*,
        Jiyong~Zhang, 
        and Chenggang Yan

\thanks{L. Bao, X. Zhou, and J. Zhang are with the School of Automation, Hangzhou Dianzi University, Hangzhou 310018, China (e-mail: lxbao@hdu.edu.cn; zxforchid@outlook.com; jzhang@hdu.edu.cn).}

\thanks{X. Lu is with the School of Software, Shandong University, Jinan 250101, China (e-mail: carrierlxk@gmail.com).}

\thanks{C. Yan is with Lishui Institute of Hangzhou Dianzi University, and School of Communication Engineering, Hangzhou Dianzi University, Hangzhou 310018, China (e-mail: cgyan@hdu.edu.cn).}

\thanks{Z. Hu is with Hangzhou Innovation Institute, Beihang University, Hangzhou 310051, China (e-mail: zhenghuihu2013@163.com).}

\thanks{Y. Sun and H. Yin are with School of Communication Engineering, Hangzhou Dianzi University and Lishui Institute of Hangzhou Dianzi University, Hangzhou 310018, China (e-mail: syq@hdu.edu.cn; yhb@hdu.edu.cn).}
\thanks{*Corresponding authors: Xiaofei Zhou, Zhenghui Hu.}


}

\maketitle

\begin{abstract}
Depth images and thermal images contain the spatial geometry information and surface temperature information, which can act as complementary information for the RGB modality. However, the quality of the depth and thermal images is often unreliable in some challenging scenarios, which will result in the performance degradation of the two-modal based salient object detection (SOD). Meanwhile, some researchers pay attention to the triple-modal SOD task, namely the visible-depth-thermal (VDT) SOD, where they attempt to explore the complementarity of the RGB image, the depth image, and the thermal image. However, existing triple-modal SOD methods fail to perceive the quality of depth maps and thermal images, which leads to performance degradation when dealing with scenes with low-quality depth and thermal images. Therefore, in this paper, we propose a quality-aware selective fusion network (QSF-Net) to conduct VDT salient object detection, which contains three subnets including the initial feature extraction subnet, the quality-aware region selection subnet, and the region-guided selective fusion subnet. Firstly, except for extracting features, the initial feature extraction subnet can generate a preliminary prediction map from each modality via a shrinkage pyramid architecture, which is equipped with the multi-scale fusion (MSF) module. Then, we design the weakly-supervised quality-aware region selection subnet to generate the quality-aware maps. Concretely, we first find the high-quality and low-quality regions by using the preliminary predictions, which further constitute the pseudo label that can be used to train this subnet. 
Finally, the region-guided selective fusion subnet purifies the initial features under the guidance of the quality-aware maps, and then fuses the triple-modal features and refines the edge details of prediction maps through the intra-modality and inter-modality attention (IIA) module and the edge refinement (ER) module, respectively. Extensive experiments are performed on VDT-2048 dataset, and the results show that our saliency model consistently outperforms 13 state-of-the-art methods with a large margin. Our code and results are available at \textcolor{blue}{https://github.com/Lx-Bao/QSFNet}.
\end{abstract}


\IEEEpeerreviewmaketitle

\section{Introduction}
\label{sec:introduction}
\IEEEPARstart{S}{alient} object detection (SOD) aims to find and segment the most attractive regions in an image. As a fundamental task in computer vision, SOD has been widely applied in many computer vision tasks, such as image/video segmentation\cite{zeng2019joint,zhou2018improving,zhou2023transformer,chen2022comprehensive}, 
defect detection\cite{zhou2021dense,wan2023lfrnet}, image quality assessment\cite{gu2016saliency}, visual tracking\cite{hong2015online}, image retrieval\cite{babenko2015aggregating}, to name a few.

\begin{figure}[!t]
\centering
\includegraphics[width=0.45\textwidth]{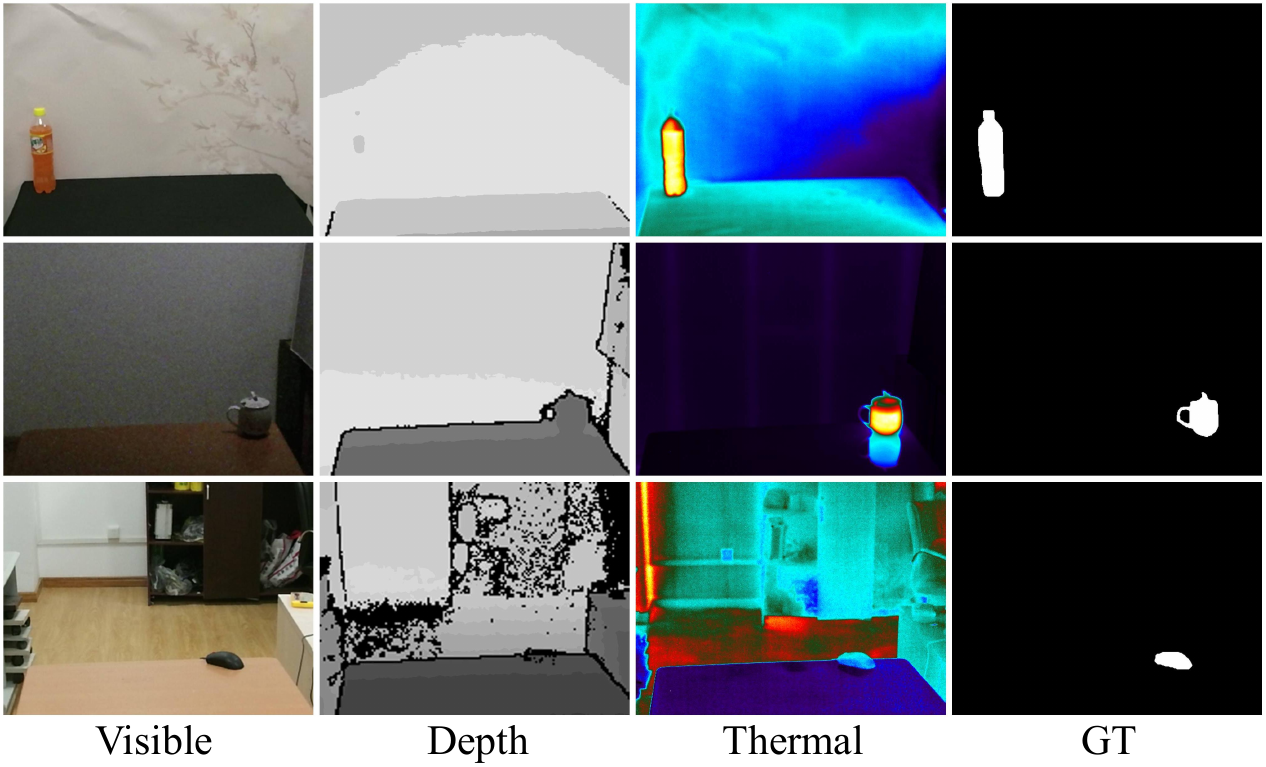}
\caption{ {Examples of VDT-2048 images: the first column shows the visible images, the second column shows the depth images, the third column shows the thermal images, and the last column presents the ground truth (GT).}}
\label{fig_examples}
\end{figure}


Although the research of SOD has achieved remarkable progress in recent years \cite{pang2020multi,zhao2019EGNet, qin2021boundary,feng2020residual,zhao2019pyramid,wang2019salient,wu2019cascaded,bao2023aggregating,song2023rethinking}, most of them still rely on appearance information such as color, contour, and texture in RGB visible images. Those approaches can achieve promising results in simple scenes, but often cannot supply satisfactory detection results when dealing with complex scenes, such as the similar appearance of background and foreground, insufficient image illumination brightness, etc. With the development of sensor technology, depth and thermal photography is successfully applied in a wide range of scenarios, especially the RGB-D and RGB-T tasks, which have attracted more and more concerns\cite{liu2021swinnet,piao2020a2dele,wen2021dynamic,chen2021depth}. Unlike RGB visible images, depth images and thermal images can provide depth-attached spatial structure cues and temperature-attached thermal radiation information, respectively, which can be regarded as complementary information for RGB images. In the past few years, a number of methods targeting RGB-D SOD \cite{fan2020bbs,liu2021tritransnet,chen2020dpanet,chen20223,cong2022cir,song2022improving} and RGB-T SOD \cite{huo2021efficient,zhou2023lsnet,tu2022weakly,zhou2023wavenet} have been designed, which have achieved promising performance.

\begin{figure}[!t]
\centering
\includegraphics[width=0.45\textwidth]{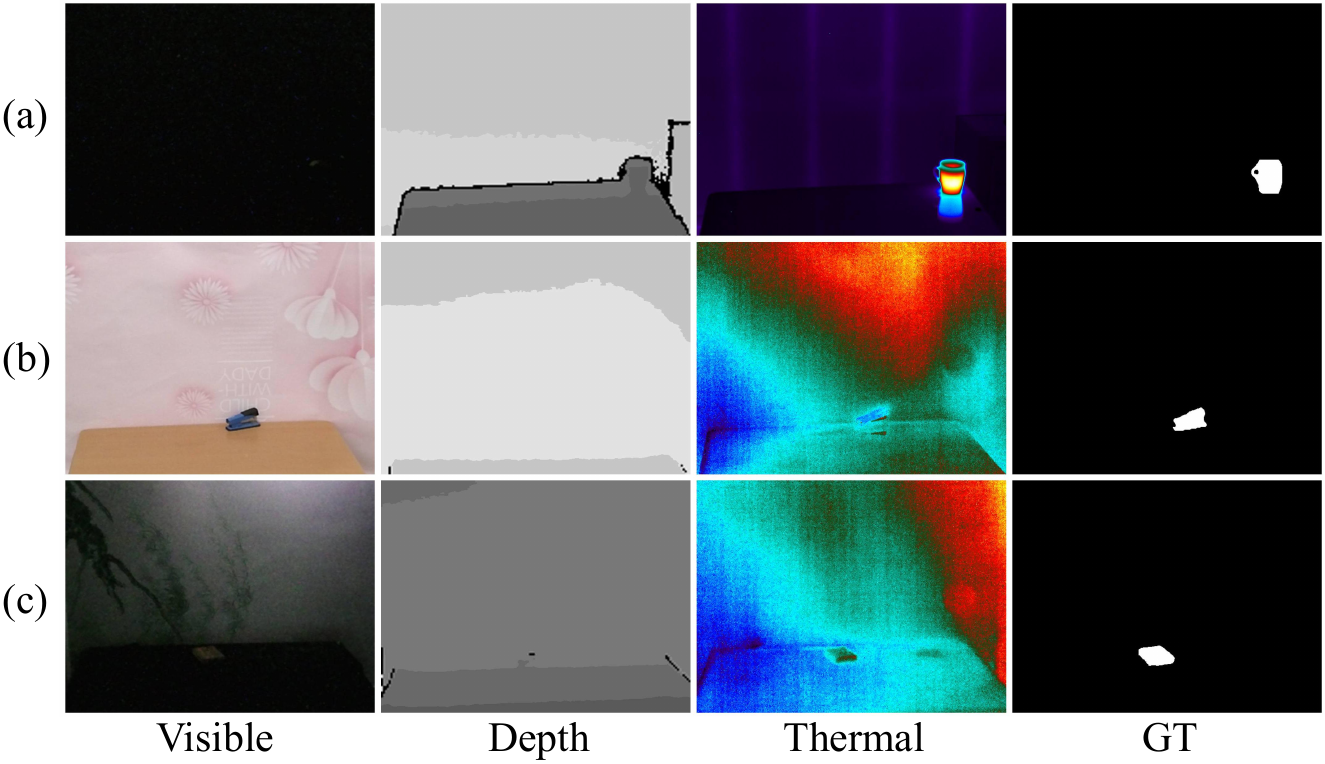}
\caption{ {Examples of some low-quality images in VDT-2048 dataset: the first column shows the visible images, and the last column presents the ground truth (GT).}}
\label{fig_challenges}
\end{figure}

 {However, depth images and thermal images have their respective drawbacks. For depth images, their efficacy is reduced in scenarios such as low illumination or with small-sized objects. For thermal images, although they are capable of depicting the spatial distribution of temperatures even in no-illumination scenes, they lack color information and texture details.} Existing SOD models mainly adopt only one modality, namely the depth image or the thermal image, to conduct salient object detection. There are fewer efforts have been devoted to the triple-modal SOD task, where they utilize the RGB images, depth images, and thermal images, simultaneously. To explore the visual perception of salient objects in a home service robot grasping task, Song \emph{et al.} \cite{song2022novel} constructed the first triple-modal SOD dataset (VDT-2048), as shown in Fig.~\ref{fig_examples}. The triple-modal images allow us to simultaneously utilize the appearance information from RGB images, the depth information from depth images, and the thermal radiation information from thermal images to obtain more comprehensive information, which can give an effective characterization of salient objects. However, compared with RGB images, depth images and thermal images are more susceptible to environmental factors, which results in their low quality in some images, as shown in Fig.~\ref{fig_challenges} (b) and (c). Besides, due to the differences in intrinsic properties, the concatenation or summation of the triple-modal information will inevitably introduce interference cues, which can significantly degrade the detection performance. Meanwhile, the existing triple-modal SOD efforts\cite{song2022novel,wan2023mffnet} employ various fusion methods (\emph{e.g.}, fully connected layer, convolutional layer, and attention mechanism) to acquire the fusion weights for the triple-modal features, but the weight often doesn’t reflect the image quality. In this way, they cannot give an effective fusion for the triple-modal information. Therefore, it is a natural idea to aggregate the triple-modal features based on the modal quality, where the quality perception can help to highlight the important information and filter out some distracting information in different-modal features.

\begin{figure*}
\centering
\includegraphics[width=0.95\textwidth]{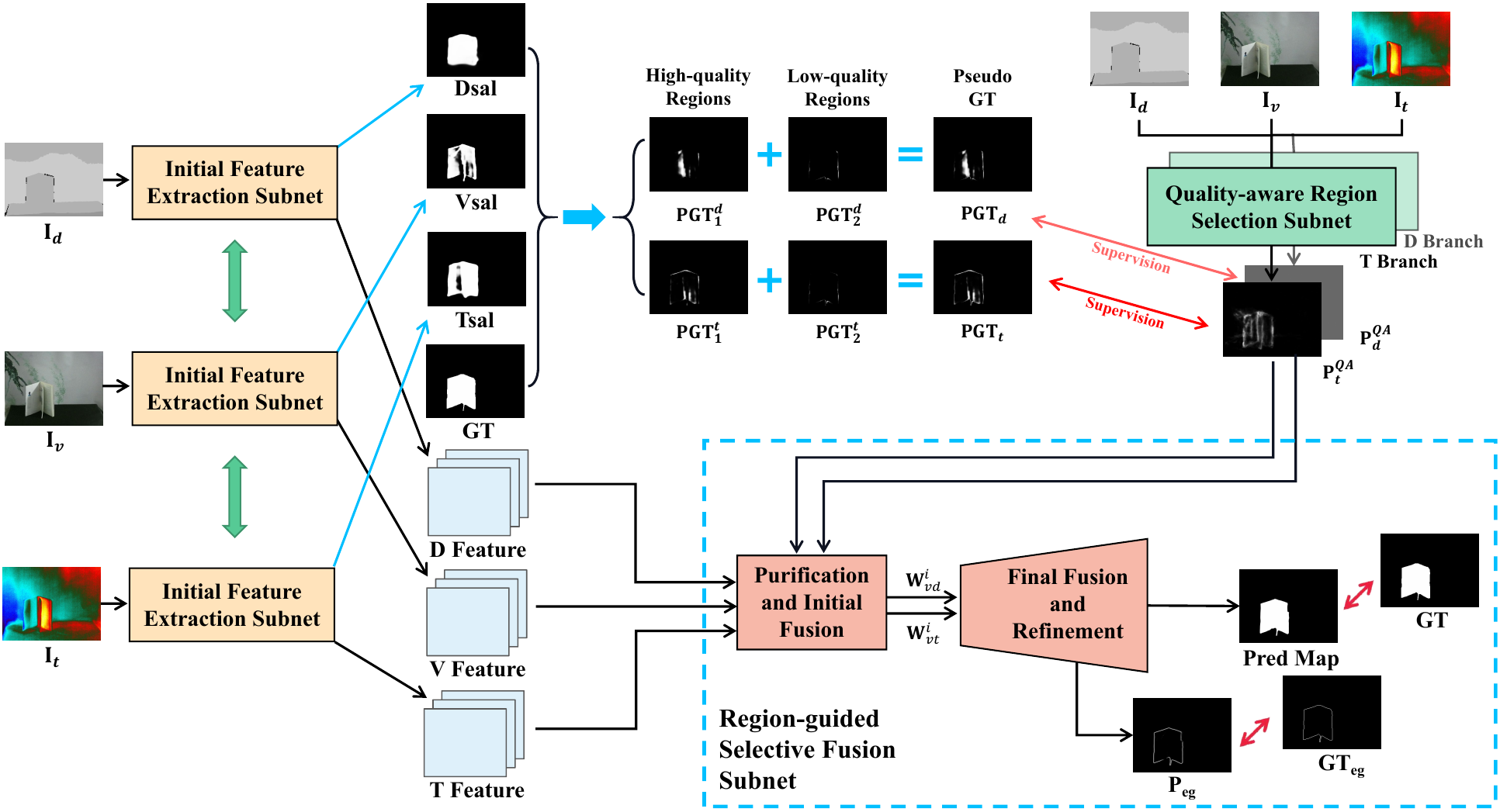}
\caption{ {The overall architecture of the proposed QSF-Net, which consists of three components including the initial feature extraction subnet, the quality-aware region selection subnet, and the region-guided selective fusion subnet. Firstly, the initial feature extraction subnet will preliminarily extract the encoder features and generate the initial prediction maps from the three modalities. Secondly, the quality-aware region selection subnet picks out the high-quality and low-quality regions, generating the quality-aware maps in a weakly-supervised way. Finally, the region-guided selective fusion subnet is deployed to purify the multi-modal features and selectively fuse the features under the guidance of the region information in the quality-aware maps.}}
\label{fig_overall}
\end{figure*}

Motivated by the aforementioned descriptions, in this paper, we propose a Quality-aware Selective Fusion Network (QSF-Net), which contains three subnets including initial feature extraction subnet, quality-aware region selection subnet, and region-guided selective fusion subnet, as presented in Fig. \ref{fig_overall}. Specifically, firstly, the initial feature extraction subnet will preliminarily extract the encoder features and generate the initial prediction maps from the three modalities. Here, we design a shrinkage pyramid architecture with a multi-scale fusion (MSF) module to extract and aggregate multi-scale features, which progressively aggregates the features of neighboring layers.

Next, we propose a quality-aware region selection subnet to pick out the high-quality and low-quality regions, which can be trained in a weakly supervised way. Generally, the high-quality regions are actually a subset of salient regions, which should be highlighted, and the low-quality regions belong to background regions, which should be suppressed. For the high-quality region of the current modality, its saliency prediction results are different from that of the other two modalities.  {That is to say, among the three prediction results of the three modalities, only the high-quality region of the current modality is correctly predicted.} Our goal is to look for such regions, which will be highlighted in the fusion process. Then, the low-quality region is where the saliency predictions of all the modalities are false. Without extra interventions, it will provide a greatly negative impact on the model performance. Consequently, we aim to find the low-quality regions, and further suppress them in the fusion process. Therefore, embarking on the selected high-quality and low-quality regions, we can acquire the pseudo ground truth, where this subnet can be trained in a weakly-supervised way. Finally, by using this subnet, we can obtain the quality-aware maps. 

After the training of the quality-aware region selection subnet, the region-guided selective fusion subnet is deployed to purify the multi-modal features and selectively fuse the features by making full use of the region information in the quality-aware maps, which gives the localization for the high-quality and low-quality regions in the multi-modal features. Besides, to conduct an effective multi-modal fusion, we design an intra-modality and inter-modality attention (IIA) module to achieve intra-modality enhancement and inter-modality interaction via self-attention and cross-attention, respectively. Moreover, we design an edge refinement (ER) module to refine the boundary details of the final fused features, which is beneficial to generating the final high-quality saliency maps.

In summary, the main contributions of this paper can be summarized as follows:\begin{enumerate}[leftmargin=*]
\item  {We propose a quality-aware selective fusion network (QSF-Net) for VDT SOD, which contains the initial feature extraction subnet, the quality-aware region selection subnet, and the region-guided selective fusion subnet. Extensive experiments are conducted on VDT-2048 dataset, which shows that the proposed QSF-Net consistently achieves the best performance when compared with the 13 state-of-the-art methods.}

\item  {We propose the initial feature extraction subnet to dig the multi-scale contextual information and acquire the detail information, where we design a shrinkage pyramid architecture with MSF modules to aggregate neighboring layer features. Particularly, we can obtain the initial prediction maps from the three modalities, which will be used to guide the generation of pseudo ground truth for the quality-aware region selection subnet.}

\item  {We design a weakly-supervised quality-aware region selection subnet to perceive quality-aware maps of depth and thermal modalities. Here, we acquire the high-quality regions and low-quality regions to generate the pseudo ground truth, which is used to train and steer the subnet to perceive the high/low-quality regions.}

\item  {We propose the region-guided selective fusion subnet to dig more comprehensive fusion information under the guidance of the quality-aware maps, where we design the intra-modality and inter-modality attention (IIA) module to dig the intra-modal and inter-modal relationships and deploy the edge refinement (ER) module to refine the edge of prediction maps.}



\end{enumerate}

The remaining of this paper is organized as follows. The related works are reviewed in Section~\ref{sec:related works}. Section~\ref{sec:proposed method} gives a detailed description of the proposed QSF-Net. In Section~\ref{sec:experiment}, comprehensive experiments and the detailed analysis are presented. Finally, the conclusion is detailed in Section~\ref{sec:conclusion}.

\section{Related Works}
\label{sec:related works}
\subsection{Single-modal Salient Object Detection}
The single-modal SOD methods use only RGB visible images to obtain the saliency map.
In the early efforts\cite{cheng2014global,zhou2017salient,li2015robust,peng2016salient,yuan2017reversion}, researchers often employ hand-crafted features to conduct saliency detection or deploy traditional machine learning algorithms to build the saliency computation model. However, due to the limitation of hand-crafted features, these methods are incapable of dealing with complex scenes.

With the rapid development of deep learning technologies, the CNN-based single-modal SOD models\cite{zhao2019EGNet,CVPR2020_LDF,qin2021boundary,feng2020residual,qin2020u2,liu2019simple,chen2020global,zhou2020interactive,pang2020multi,li2020stacked} outperform the traditional saliency methods.
For example,
Pang \emph{et al.} \cite{pang2020multi} and Huang \emph{et al.}\cite{huang2019saliency} explored the interaction of multi-level and multi-scale information. Zhao \emph{et al.}\cite{zhao2019pyramid} and Chen \emph{et al.}\cite{chen2018reverse} attempted to deploy visual attention mechanisms to enhance features and filter-out backgrounds. Chen \emph{et al.} \cite{chen2020global} and Liu \emph{et al.}\cite{liu2019simple} tried to aggregate the global context-aware information and the low-level features. In addition, the deployment of edge information is a crucial factor for capturing intricate details within boundary regions. For example, Zhao \emph{et al.}\cite{zhao2019EGNet} and Zhou \emph{et al.} \cite{zhou2020interactive} explored the complementarity between features of salient objects and features of salient edge, yielding the high-quality saliency maps. Zhou \emph{et al.}\cite{zhou2021edge,zhou2022edge} fully utilized edge cues to provide precise boundary details to enhance the deep features in both explicit and implicit ways.

\subsection{Two-modal Salient Object Detection}
Depth information and thermal information are effective for SOD, and they can provide rich spatial distance information and thermal radiation information, which are the complementary information to the RGB cues. For the depth information,
Fan \emph{et al.}\cite{fan2020bbs} introduced a depth-enhanced module to mine abundant spatial cues in depth images, and designed a bifurcated backbone strategy to guide the decoding process. Piao \emph{et al.}\cite{piao2020a2dele} designed a depth distiller to transfer the depth knowledge from the depth stream to the RGB stream, which ensures the reliable transmission of depth information by eliminating redundant depth cues. Liu \emph{et al.}\cite{liu2021tritransnet} proposed a triplet transformer embedding network to enhance the fused high-level semantic features, and utilized the enhanced semantic features to guide the decoding process.  {Chen \emph{et al.} \cite{chen2020improved} proposed a two-stage depth estimation strategy to refine depth information, and presented a selective deep fusion network to integrate high-quality depth information with color information. Wang \emph{et al.} \cite{wang2021depth} introduced a depth quality-aware selective fusion strategy, which leverages multi-scale features to assess depth quality and guide the integration of RGB and depth information.}

Meanwhile, in some extremely complex environments (\emph{e.g.}, low illumination, occlusion, fog, haze, etc.), thermal images can overcome the above challenges. Wang \emph{et al.}\cite{wang2021cgfnet} designed a variety of fusion guidance modules in decoder to provide multi-scale global semantic information and integrate multi-level features, which fuses the cross-modality information in a more effective way. Huo \emph{et al.}\cite{huo2021efficient} proposed a fusion module to explore the complementary cues between RGB modality and T modality, and also designed a stacked refinement network to refine the segmentation results. Zhou \emph{et al.}\cite{zhou2023lsnet} introduced a spatial boosting network for efficient RGB-T SOD, where they utilized a boundary boosting algorithm to refine the predicted saliency maps.

Recently, a growing number of saliency methods have been proposed to explore the similarity between RGB-D SOD and RGB-T SOD tasks, which supply impressive performance on both two tasks. For example, Liu \emph{et al.}\cite{liu2021swinnet} utilized the powerful feature extraction capabilities of Swin Transformer to obtain hierarchical features of each modality. Besides, they optimized and fused intra-layer cross modality through a spatial alignment and channel re-calibration module. Tang \emph{et al.}\cite{tang2022hrtransnet} inserted a supplementary modality into the primary modality, which enables the fusion of two-modal information at the input level. Chen \emph{et al.}\cite{chen2022modality} employed the structural similarity between two modalities to guide the fusion of the multi-modal information, which effectively reduces the semantic gap.

\subsection{Triple-modal Salient Object Detection}

Despite numerous methods have been proposed for the two-modal SOD, the research on triple-modal SOD is still in its early stages. Song \emph{et al.}\cite{song2022novel} constructed the first triple-modal SOD dataset (VDT-2048), where visual perception in grasping tasks is explored for home service robots. Besides, they proposed a novel triple-modal SOD method, namely HWSI, which first achieved the interactive fusion of two modalities through the attention mechanism and then weighted the two-modal features to accomplish the complementary aggregation of triple-modal information. Wan \emph{et al.}\cite{wan2023mffnet} proposed a multi-modal feature fusion network, which excavated the complementarity of the three modalities in the encoder and conducted progressive feature decoding by aggregating contextual features, edge cues, and multi-scale information.

The above researchers have explored the complementarity during the triple-modal fusion step. However, they often neglect that depth and thermal images are highly susceptible to environmental factors, which will lead to the poor overall quality of the two types of images. The complementary information provided by the two modalities may be incorrect, and it is harmful to the fusion of triple-modal cues. Therefore, we propose a modality quality-aware selective fusion network, which selects regions in each modality according to the quality perception results.

\section{The Proposed Method}
\label{sec:proposed method}
In this section, we detail the proposed QSF-Net. In Section~\ref{sec:Network Overview}, we give an overview of the proposed QSF-Net. In Section ~\ref{sec:initial feature extraction subnet}, we present the initial feature extraction subnet of triple-modality. After that, we detail the quality-aware region selection subnet presented in Section~\ref{sec:quality-aware region selection subnet} and the region-guided selective fusion subnet presented in Section~\ref{sec:region-guided selective fusion subnet}. Besides, in Section~\ref{sec:loss function}, we clarify the training process and loss function of our network.

\subsection{Network Overview}
\label{sec:Network Overview}
As shown in Fig.~\ref{fig_overall}, our proposed QSF-Net consists of three subnets, \emph{i.e.}, initial feature extraction subnet, quality-aware region selection subnet, and region-guided selective fusion subnet. Firstly, in the initial feature extraction subnet, the input visible, depth, and thermal images $\mathbf{I}_v$, $\mathbf{I}_d$, $\mathbf{I}_t$ are delivered into the shared-weighted encoder, and then the decoding process of the three modalities is independently executed, generating the initial features and initial saliency prediction maps of each modality, respectively. Particularly, to effectively aggregate multi-scale features, the decoder employs a feature shrinkage pyramid structure with multi-scale fusion (MSF) modules. Next, based on the initial prediction maps, we can generate the pseudo ground truth for the weakly supervised quality-aware region selection subnet, which can infer the location of high-quality and low-quality regions via the quality-aware maps $\mathbf{P}_{d}^{QA}$ and $\mathbf{P}_{t}^{QA}$.  After that, under the guidance of the quality-aware maps, the region-guided selective fusion subnet, which includes intra-modality and inter-modality attention (IIA) module and edge refinement (ER) module, selectively purifies and fuses triple-modal features, and then refines the boundary details of the fused features, yielding the final high-quality saliency map.

\begin{figure*}
\centering
\includegraphics[width=0.92\textwidth]{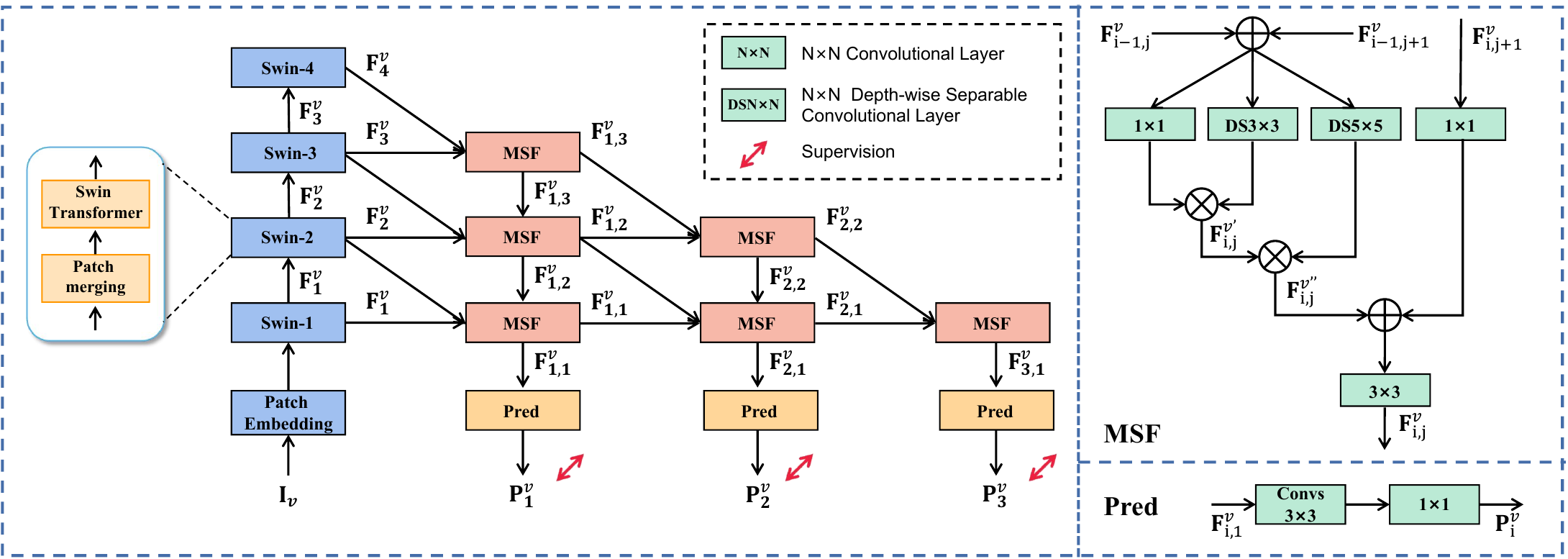}
\caption{ {The architecture of the initial feature extraction subnet (visible branch).}}
\label{fig_FIENet_sig}
\end{figure*}

\begin{figure}[!t]
\centering
\includegraphics[width=0.45\textwidth]{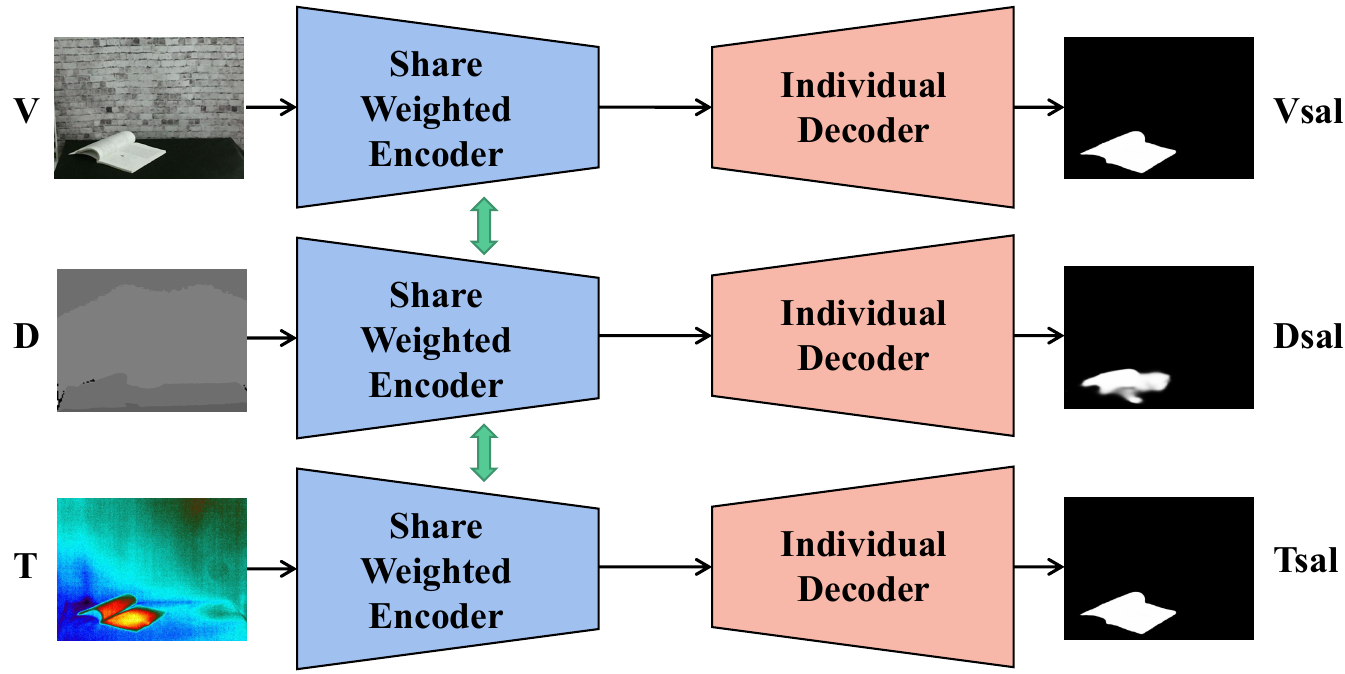}
\caption{The overall architecture of the initial feature extraction subnet.}
\label{fig_FIENet_all}
\end{figure}

\subsection{Initial Feature Extraction Subnet}
\label{sec:initial feature extraction subnet}

As depicted in Fig.~\ref{fig_FIENet_all}, our initial feature extraction subnet adopts an encoder-decoder architecture, where the encoder of each branch is built on the Swin Transformer\cite{liu2021swin} and the three encoders are equipped with the shared weight strategy. In this way, we can extract long-range dependency features.  {Formally, the inputs of our network are three-modality images, namely $\mathbf{I}_v$, $\mathbf{I}_d$, $\mathbf{I}_t$ $\in \mathbb{R}^{H\times W \times3}$. Through the encoder, we can obtain three groups of features, namely $\left\{\mathbf{F}_i^v\right\}_{i=1}^4$, $\left\{\mathbf{F}_i^d\right\}_{i=1}^4$, and $\left\{\mathbf{F}_i^t\right\}_{i=1}^4$.}
However, in Transformer, the self-attention is computed on the patch partition, which leads to the insufficient representation of local details. To solve this problem, the decoder adopts a feature shrinkage pyramid structure with multi-scale fusion (MSF) modules, as shown in the left part of Fig. \ref{fig_FIENet_sig}. Taking the visible branch as an example, we can see that the decoder aggregates adjacent-level Transformer features via a layer-by-layer shrinkage pyramid structure, and in this way, we can acquire effective decoder features.  {Meanwhile, to better explore the multi-scale contextual information of the decoder features, we designed the MSF module, which utilizes convolution layers with different kernel sizes to extend the receptive fields.  As shown in the right part of Fig.~\ref{fig_FIENet_sig}, the entire process is formulated as follows:}


\begin{equation}\label{eq1}
\left \{\begin{array}{l}
\widetilde{\mathbf{F}}_{i,j}^v= \mathbf{F}_{i,j}^v + \mathbf{F}_{i,j+1}^v \vspace{2ex}\\
\mathbf{F}_{i,j}^{v'}=\operatorname{Conv}_{1\times1}(\widetilde{\mathbf{F}}_{i,j}^v) \times \operatorname{DSConv}_{3\times3}(\widetilde{\mathbf{F}}_{i,j}^v)  \vspace{2ex}\\
\mathbf{F}_{i,j}^{v''}=\mathbf{F}_{i,j}^{v'} \times \operatorname{DSConv}_{5\times5}(\widetilde{\mathbf{F}}_{i,j}^v)  \vspace{2ex}\\
\mathbf{F}_{i,j}^{v}=\operatorname{Conv}_{3\times3}\big(\mathbf{F}_{i,j}^{v''} + \operatorname{Conv}_{1\times1}(\mathbf{F}_{i,j+1}^v)\big) \end{array},
\right.
\end{equation}
where $\operatorname{Conv}_{1\times1}(\cdot)$ and $\operatorname{Conv}_{3\times3}(\cdot)$  represents $1\times1$ and $3\times3$ convolutional layers, respectively. $\operatorname{DSConv}_{3\times3}(\cdot)$ and $\operatorname{DSConv}_{5\times5}(\cdot)$ means $3\times3$ and $5\times5$ depth-wise separable convolution layer \cite{chollet2017xception}, respectively. $\times/+$ is element-wise multiplication/summation.


After that, the aggregated features are delivered to the next layer and the next stage. Finally, we can obtain the initial features $\left\{\mathbf{F}_{i,1}^{v}\right\}_{i=1}^3$, which are the outputs of the bottom MSF modules. We further pass the feature into the prediction (Pred) module to generate the initial prediction map $\left\{\mathbf{P}_i^v\right\}_{i=1}^3$. This process can be written as follows:
\begin{equation}\label{eq2}
\mathbf{P}_{i}^{v}=\operatorname{Conv}_{1\times1}\big(\operatorname{Convs}_{3\times3}(\mathbf{F}_{i,1}^{v})\big),
\end{equation}
where the Pred module contains a 1$\times$1 convolutional layer, a 3$\times$3 convolutional layer, a batch normalization (BN) layer, and a ReLU activation function, namely the function $\operatorname{Conv}_{1\times1}(\operatorname{Convs}_{3\times3}(\cdot))$. Here, $\operatorname{Convs}_{3\times3}(\cdot)$ represents $3\times3$ convolutional layers followed by a BN layer and a ReLU activation function.

\begin{figure*}
\centering
\includegraphics[width=0.98\textwidth]{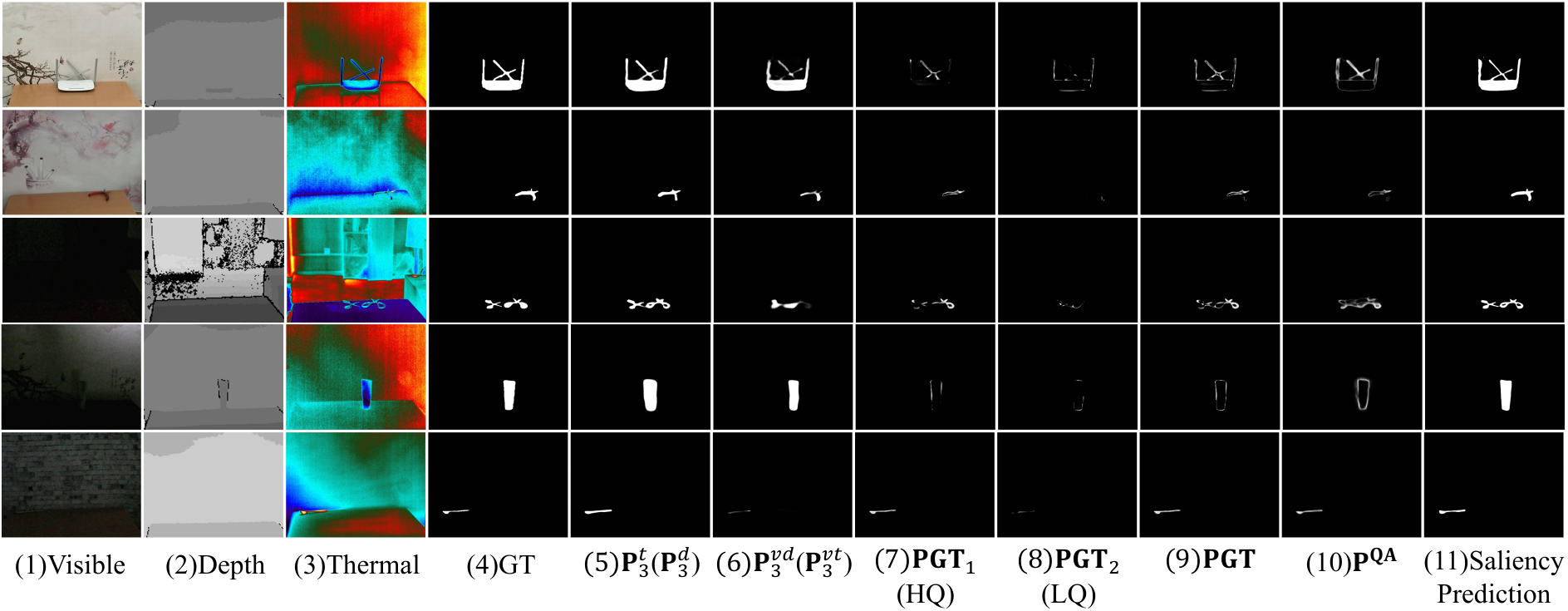}
\caption{ {The visualization of the results in the quality-aware region selection subnet and the region-guided selective fusion subnet. The $1^{st}$-$4^{th}$ columns show the visible, depth, thermal images and ground truth (GT), respectively. The $5^{th}$ column shows the initial prediction map of thermal or depth modality, namely $\mathbf{P}^{t}_3$ and $\mathbf{P}^{d}_3$. In the $6^{th}$ column, $\mathbf{P}^{vd}_3$ means the average saliency value of the RGB and depth saliency predictions and $\mathbf{P}^{vt}_3$ means the average saliency value of the RGB and thermal saliency predictions. The $7^{th}$ and $8^{th}$ columns show the high-quality and low-quality regions, respectively. $\mathbf{PGT}$ ($9^{th}$ column) is the pseudo GT and $\mathbf{P}^{QA}$ ($10^{th}$ column) is the quality-aware map. The $11^{th}$ column shows the final saliency prediction under the guidance of quality-aware map $\mathbf{P}^{QA}$.}}
\label{fig_MV}
\end{figure*}

\begin{figure}
\centering
\includegraphics[width=0.45\textwidth]{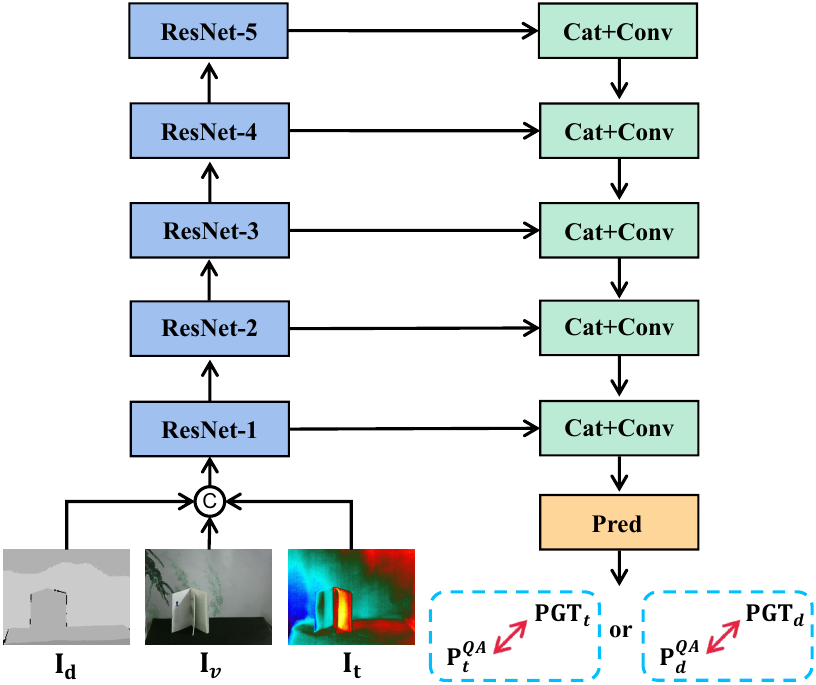}
\caption{The architecture of the quality-aware region selection subnet.}
\label{fig_QA}
\end{figure}

\subsection{Quality-aware Region Selection Subnet}
\label{sec:quality-aware region selection subnet}
Compared with RGB images, the overall quality of depth images and thermal images is lower.  {Meanwhile, we find that the two-modality images contain complementary cues, especially when dealing with some challenging scenes including low illumination, low contrast between salient regions and backgrounds, and cluttered backgrounds.} Therefore, to achieve the complementary fusion with RGB features, we design the quality-aware region selection subset to identify the high-quality and low-quality regions from depth and thermal features, yielding the quality-aware map under the supervision of selected regions. Actually, the high-quality regions and the low-quality regions belong to the salient regions and background regions, respectively. Here, we take the depth branch as an example. There are two points for the selection of regions. Firstly, we pick out high-quality regions that are correctly detected in the depth modality but falsely suppressed in the visible and thermal modalities. Under this condition, the highlight of this region in the fusion stage will be in a dilemma. Secondly, we select low-quality regions, where the prediction of backgrounds in depth, visible, and thermal modalities are all falsely highlighted as salient regions.

Specifically, the prediction maps of visible, depth, and thermal modalities as well as the ground truth ($\mathbf{GT}$) are available in the training of initial feature extraction subnet, and thus we attempt to generate the pseudo GT for the training of quality-aware region selection subnet. Firstly, according to the first point, we compute the pseudo GT $\mathbf{PGT}_{1}^{d}$, namely the high-quality region, which can be written as
\begin{equation}\label{eq3}
\mathbf{PGT}_{1}^{d}=\operatorname{ReLU}\big(\mathbf{P}_{3}^{d}-(\mathbf{P}_{3}^{v}+\mathbf{P}_{3}^{t})/2\big)\times \mathbf{GT},
\end{equation}
 {where $\operatorname{ReLU}(\cdot)$ means ReLU activation function. Here, for the salient regions highlighted by $\mathbf{GT}$, we aim to locate the high-quality regions, where these regions' depth saliency prediction $\mathbf{P}_3^d$ are larger than the average saliency value of the RGB and thermal saliency predictions (\emph{i.e.,} $\mathbf{P}^{v}_{3}$ and $\mathbf{P}^{t}_{3}$). This indicates that such regions can be better highlighted by depth images instead of the RGB and thermal counterparts.}

Then, according to the second point, we compute the pseudo GT $\mathbf{PGT}_{2}^{d}$, namely the low-quality regions, which can be formulated as
\begin{equation}\label{eq4}
\mathbf{PGT}_{2}^{d}=\big(\mathbf{P}_{3}^{d}\times(\mathbf{P}_{3}^{v}+\mathbf{P}_{3}^{t})/2\big)\times (1-\mathbf{GT}).
\end{equation}
 {Here, for the background regions indicated by the GT, we attempt to locate the low-quality regions, where the depth, RGB, and thermal modalities jointly treat the backgrounds as saliency regions.} Meanwhile, we should note that by using the Eq. \ref{eq4}, the low-quality regions are highlighted.

Finally, we combine the two regions to get our complete pseudo GT $\mathbf{PGT}_{d}$ for the depth modality, which is written as
\begin{equation}\label{eq5}
\mathbf{PGT}_{d}=\mathbf{PGT}_{1}^{d}+\mathbf{PGT}_{2}^{d}.
\end{equation}

Following the aforementioned steps, we can acquire the pseudo GT including $\mathbf{PGT}_d$ and $\mathbf{PGT}_t$ for the depth modality and thermal modality, respectively. After that, we use the pseudo GT to supervise the training of our quality-aware region selection subnet. Here, we also take the depth modality as an example. Specifically, as shown in Fig.~\ref{fig_QA}, we adopt an encoder-decoder architecture, where the encoder is built on the ResNet-34\cite{he2016deep}. Here, we should note that the input of this subnet is the concatenation of three-modality images, namely $\mathbf{I}_v$, $\mathbf{I}_d$, and $\mathbf{I}_t$, along the channel dimension.  {The encoder can generate five-level features, and the decoder progressively integrates each-level decoder feature and encoder feature, yielding the quality-aware map $\mathbf{P}_{d}^{QA}$.} In the same way, we can also obtain the quality-aware map $\mathbf{P}_{t}^{QA}$ for the thermal modality.  {And we have demonstrated the qualitative results of quality-aware maps in Fig.~\ref{fig_MV}.}

\begin{figure}
\centering
\includegraphics[width=0.45\textwidth]{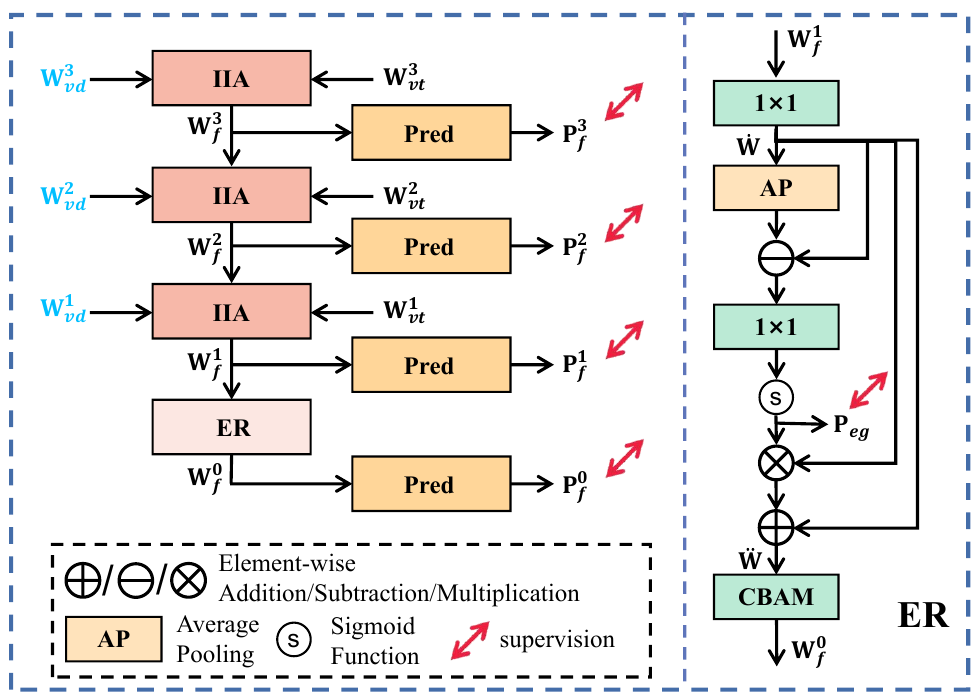}
\caption{ {The architecture of the final fusion and refinement stage of region-guided selective fusion subnet and the ER module.}}
\label{fig_SF}
\end{figure}

\subsection{Region-guided Selective Fusion Subnet}
\label{sec:region-guided selective fusion subnet}
\subsubsection{ {Purification and Initial Fusion Stage}}
Embarking on the quality-aware maps including $\mathbf{P}_{d}^{QA}$ and $\mathbf{P}_{t}^{QA}$, we can purify the initial features including $\left\{\mathbf{F}_{i,1}^{v}\right\}_{i=1}^3$, $\left\{\mathbf{F}_{i,1}^{d}\right\}_{i=1}^3$, and $\left\{\mathbf{F}_{i,1}^{t}\right\}_{i=1}^3$. Specifically, taking the V-D modality (\emph{i.e.}, visible and depth) as an example. First, following the generation of high-quality region $\mathbf{PGT}_1^d$, namely Eq.~\ref{eq3}, we introduce quality-aware map $\mathbf{P}_{d}^{QA}$ to generate the purified visible-depth features $\mathbf{W}_{vd}^{i,1}$, which should be emphasized in fusion step. The process can be written as
\begin{equation}\label{eq6}
\mathbf{W}_{vd}^{i,1}=\big(\mathbf{F}_{i,1}^{d}-(\mathbf{F}_{i,1}^{v}+\mathbf{F}_{i,1}^{t})/2\big) \times \mathbf{P}_{d}^{QA}.
\end{equation}

Then, we eliminate the low-quality regions in the initial RGB features to suppress the interference information, generating the purified visible-depth feature $\mathbf{W}_{vd}^{i,2}$. The process can be defined as
\begin{equation}\label{eq7}
\mathbf{W}_{vd}^{i,2}=\mathbf{F}_{i,1}^{v} \times (1-\mathbf{P}_{d}^{QA}).
\end{equation}
Here, according to Eq. \ref{eq3}, there are no high-quality regions in $\mathbf{F}_{i,1}^{v}$, and by reversing the $\mathbf{P}_{d}^{QA}$, we can set the value of low-quality regions to 0. In this way, we can filter out the low-quality regions from $\mathbf{F}_{i,1}^{v}$.

Finally, the purified visible-depth features including $\mathbf{W}_{vd}^{i,1}$ and $\mathbf{W}_{vd}^{i,2}$ are combined by element-wise addition, obtaining the initial fused feature $\mathbf{W}_{vd}^{i}$, namely
\begin{equation}
\mathbf{W}_{vd}^{i}=\mathbf{W}_{vd}^{i,1}+\mathbf{W}_{vd}^{i,2}.
\end{equation} Following the aforementioned steps, we can acquire the initial fused features $\left\{\mathbf{W}_{vd}^{i}\right\}_{i=1}^3$ for the visible-depth modality.  Similarly, we can also obtain the initial fused feature $\left\{\mathbf{W}_{vt}^{i}\right\}_{i=1}^3$ for the visible-thermal modality.

\subsubsection{ {Final Fusion and Refinement Stage}}
 {After that, we deploy the final fusion and refinement stage of region-guided selective fusion subnet to aggregate the two groups of features,} generating the final high-quality saliency map $\mathbf{P}_f^0$, as shown in Fig.~\ref{fig_SF}.
Overall, the inputs of the subnet are the two groups of features including $\left\{\mathbf{W}_{vd}^{i}\right\}_{i=1}^3$ and $\left\{\mathbf{W}_{vt}^{i}\right\}_{i=1}^3$.  {Firstly, for each-level features $\mathbf{W}_{vd}^{i}$ and $\mathbf{W}_{vt}^{i}$, we deploy the IIA module to enhance and interact them, and then combine them with the output feature of the previous IIA module, which can generate the fused feature $\mathbf{W}_{f}^{i}$.  
The entire process can be formulated as follows,}
\begin{equation}\label{eq9}
\left \{\begin{array}{l}
\mathbf{W}_{f}^{i} = \operatorname{IIA}(\mathbf{W}_{vd}^{i},\mathbf{W}_{vt}^{i}) \quad i=1 \vspace{2ex}\\
\mathbf{W}_{f}^{i} = \operatorname{IIA}(\mathbf{W}_{vd}^{i},\mathbf{W}_{vt}^{i},\mathbf{W}_{f}^{i-1}) \quad i=2,3
\end{array}.
\right.
\end{equation}

In this way, we can progressively obtain three fused features $\left\{\mathbf{W}_{f}^{i}\right\}_{i=1}^3$. After that, we employ an edge refinement (ER) module to emphasize the spatial details of the last fused feature $\mathbf{W}_{f}^{1}$, generating the final high-quality output feature $\mathbf{W}_{f}^{0}$, namely
\begin{equation}\label{eq10}
\mathbf{W}_{f}^{0} = \operatorname{ER}(\mathbf{W}_{f}^{1}).
\end{equation}
Besides, we deploy four prediction (Pred) modules on the four fused features $\left\{\mathbf{W}_{f}^{i}\right\}_{i=0}^3$  to generate the saliency predictions $\left\{\mathbf{P}_{f}^{i}\right\}_{i=0}^3$ for deep supervision. The process can be formulated as
\begin{equation}\label{eq2-5}
\mathbf{P}_{f}^{i}=\operatorname{Conv}_{1\times1}\big(\operatorname{Convs}_{3\times3}(\mathbf{W}_{f}^{i})\big).
\end{equation}

\begin{figure}
\centering
\includegraphics[width=0.45\textwidth]{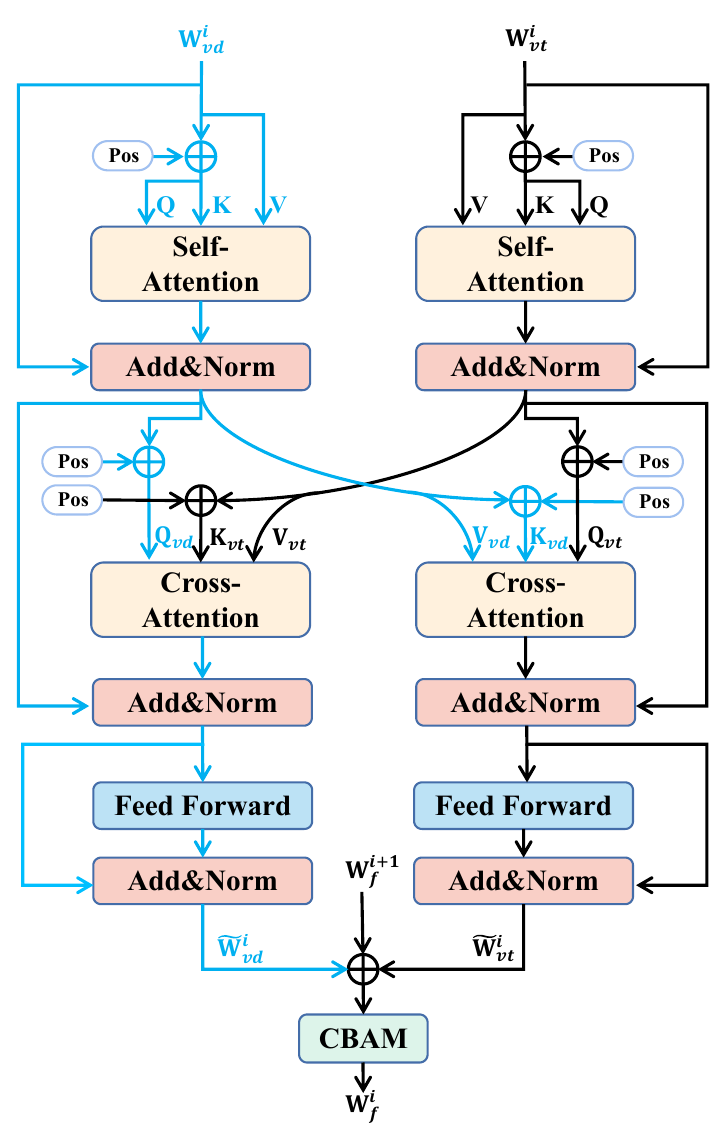}
\caption{The architecture of the Intra-modality and Inter-modality Attention (IIA) module.}
\label{fig_IIA}
\end{figure}

Here, to model the correlations within and between the fused modalities, we design an Intra-modality and Inter-modality Attention (IIA) module, as shown in Fig.~\ref{fig_IIA}. Formally, firstly, the two input features $\mathbf{W}_{vd}^{i}$ and $\mathbf{W}_{vt}^{i}$ are divided into multiple tokens, and each with a size of 1$\times$1$\times$128. Secondly, an intra-modality self-attention layer is applied to each modality, which explores the intra-modality relation with the global receptive field. Next, an inter-modality cross-attention layer is employed to explore the inter-modality relationship and establish the long-term dependency between the two modalities. The above attention layers are both implemented by efficient attention \cite{shen2021efficient}, which can reduce the memory and computational costs. Finally, after a feed-forward (FF) layer, the features of each modality $\widetilde{\mathbf{W}}_{vd}^{i}$ and $\widetilde{\mathbf{W}}_{vt}^{i}$ will be fused with the $\mathbf{W}_{f}^{i+1}$ from the previous IIA module, generating the final fused feature $\mathbf{W}_{f}^{i}$ by the CBAM module\cite{woo2018cbam}. The entire process can be written as follows,

\begin{equation}\label{eq2-5}
\left \{\begin{array}{l}
\widetilde{\mathbf{W}}_{vd}^{i} = \operatorname{FF}\Big(\operatorname{CA}\big(\operatorname{SA}(\mathbf{W}_{vd}^{i}), \mathbf{W}_{vt}^{i}\big)\Big) \vspace{2ex}\\
\widetilde{\mathbf{W}}_{vt}^{i} = \operatorname{FF}\Big(\operatorname{CA}\big(\operatorname{SA}(\mathbf{W}_{vt}^{i}), \mathbf{W}_{vd}^{i}\big)\Big) \vspace{2ex}\\
\mathbf{W}_{f}^{i} = \operatorname{CBAM}(\widetilde{\mathbf{W}}_{vd}^{i} + \widetilde{\mathbf{W}}_{vt}^{i} + \mathbf{W}_{f}^{i+1})
\end{array},
\right.
\end{equation}
where $\operatorname{SA}(\cdot)$ and $\operatorname{CA}(\cdot)$ mean self-attention and cross-attention, respectively. $\operatorname{CBAM}(\cdot)$ represents the convolutional block attention module \cite{woo2018cbam}.

Edge information is an important factor for enhancing spatial boundary details\cite{zhou2021edge,zhou2022edge}. Therefore, we propose an edge refinement (ER) module to refine edge details, as shown in the right part of Fig.~\ref{fig_SF}. Specifically, after obtaining the feature $\mathbf{W}_f^1$, we firstly extract the edge of the feature by using average pooling and subtraction operations. Then, a 1$\times$1 convolution layer with a sigmoid function is utilized to predict the edge, which is supervised by the ground truth of salient edges. After that, the predicted edge is used to emphasize the feature’s edge information via element-wise multiplication. Next, the enhanced feature is combined with the original features in a residual way, and the fused feature is further processed by a CBAM module, yielding the final output feature. The whole process can be formulated as follows,

\begin{equation}\label{eq2-5}
\left \{\begin{array}{l}
\dot{\mathbf{W}} = \operatorname{Conv}_{1\times1}(\mathbf{W}_{f}^{1})  \vspace{2ex}\\
\ddot{\mathbf{W}} = \dot{\mathbf{W}} + \dot{\mathbf{W}} \times \operatorname{\sigma}\Big(\operatorname{Convs_{1\times 1}}\big(\dot{\mathbf{W}} - \operatorname{AP}(\dot{\mathbf{W}})\big)\Big)  \vspace{2ex}\\
\mathbf{W}_f^0 = \operatorname{CBAM}(\ddot{\mathbf{W}})
\end{array},
\right.
\end{equation}
where $\operatorname{\sigma}(\cdot)$ means sigmoid function. $\operatorname{Convs_{1\times 1}}(\cdot)$ represents a 1$\times$1 convolution layer with a batch normalization layer. $\operatorname{AP}(\cdot)$ represents the average pooling operation  {(kernel size=3$\times$3, stride=1)}.

\begin{figure}
\centering
\includegraphics[width=0.48\textwidth]{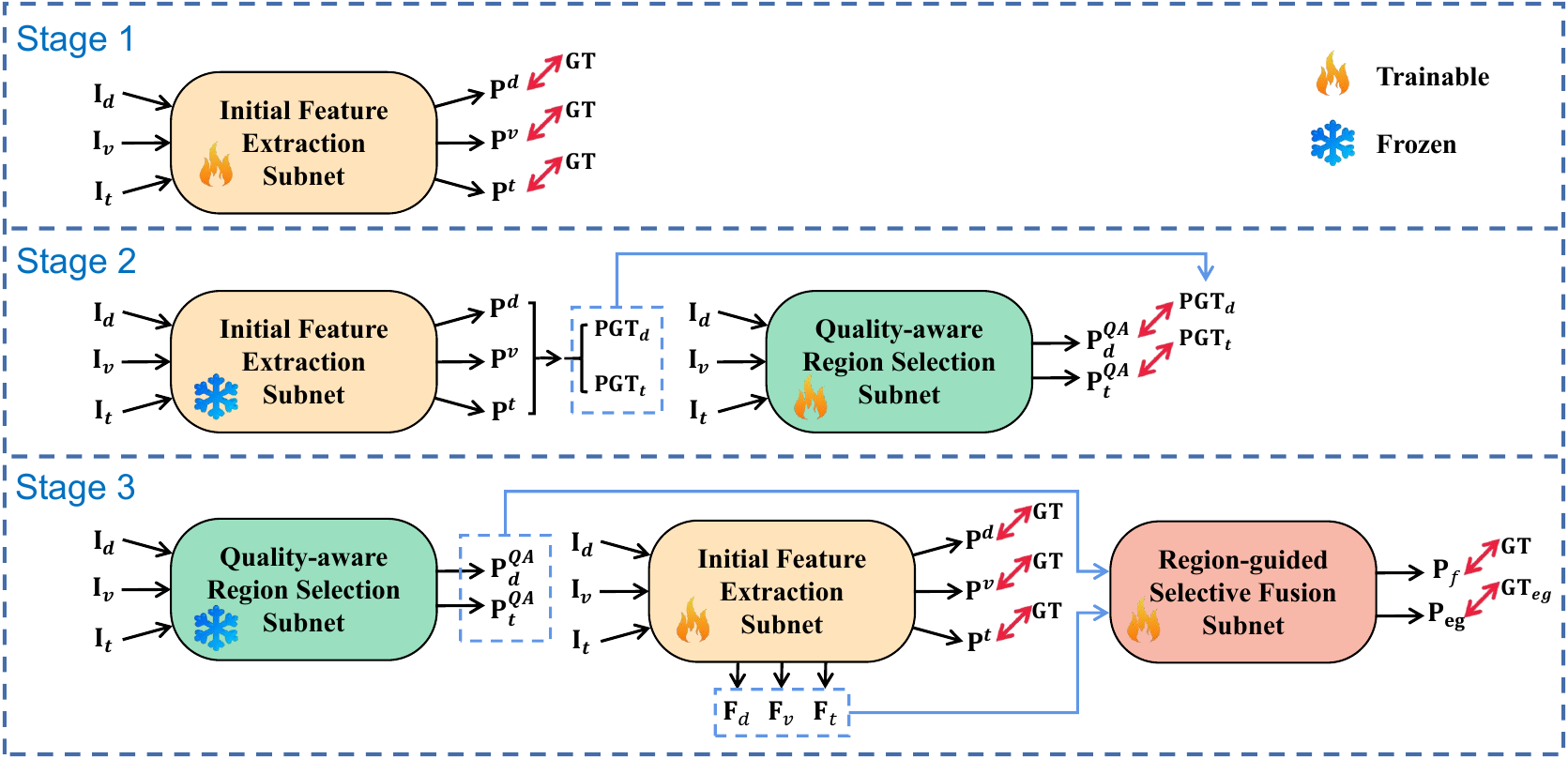}
\caption{ {The visualization of our stage-wise training process.}}
\label{fig_train}
\end{figure}

\subsection{Network Training and Loss Function}
\label{sec:loss function}
 {To summarize, our network training is divided into three stages, as presented in Fig.~\ref{fig_train}. In the first stage, we train the initial feature extraction subnet to obtain the initial features and the inital prediction maps for each modality. In the second stage, based on the initial prediction maps, we formulate the pseudo GTs (\emph{i.e.,} Eq.~\ref{eq3}, Eq.~\ref{eq4}, and Eq.~\ref{eq5}) to supervise the training of the quality-aware region selection subnet, in which the parameters of the initial feature extraction subnet are frozen.}

 {In the third stage, we combine the three stages. We first get the initial features and coarse prediction maps for each modality branch through the initial feature extraction subnet. Then, we use the quality-aware region selection subnet to perceive the most valuable regions including the high-quality regions and the low-quality regions. Finally, under the guidance of the valuable regions, we fuse the multi-modal features to generate the final high-quality prediction maps in the region-guided selective fusion subnet. In this stage, the parameters of the quality-aware region selection subnet are frozen, and the other two subnets are involved in the training process.}

Meanwhile, a suitable supervision strategy and loss function can improve the performance of the network without extra parameters. In the first stage, we adopt the pixel position aware (PPA) loss\cite{wei2020f3net}, which includes a weighted binary cross entropy (BCE) loss\cite{de2005tutorial} and a weighted IoU loss\cite{rahman2016optimizing}, to supervise the prediction maps, which can be formulated as follows,
\begin{equation}\label{eq-loss}
\left \{\begin{array}{l}
\mathcal{L}_{v}^1 =\sum_{i=1}^{3}{L_{PPA}(\mathbf{P}_{i}^{v}, \mathbf{GT})} \vspace{2ex}\\
\mathcal{L}_{d}^1 =\sum_{i=1}^{3}{L_{PPA}(\mathbf{P}_{i}^{d}, \mathbf{GT})} \vspace{2ex}\\
\mathcal{L}_{t}^1 =\sum_{i=1}^{3}{L_{PPA}(\mathbf{P}_{i}^{t}, \mathbf{GT})} \vspace{2ex}\\
\mathcal{L}_{total}^1 = \mathcal{L}_{v}^1 + \mathcal{L}_{d}^1 + \mathcal{L}_{t}^1
\end{array},
\right.
\end{equation}
where $\mathcal{L}_{total}^1$ is the total loss value of stage one, and $L_{PPA}$ means the PPA loss.

In the second stage, we supervise the final output of each modality using pseudo GT by using the BCE loss function, and thus the total loss $\mathcal{L}_{total}^2$ can written as follows
\begin{equation}\label{eq-loss}
\left \{\begin{array}{l}
\mathcal{L}_{d}^2 = L_{BCE}(\mathbf{P}_{d}^{QA},\mathbf{PGT}_{d}) \vspace{2ex}\\
\mathcal{L}_{t}^2 = L_{BCE}(\mathbf{P}_{t}^{QA},\mathbf{PGT}_{t}) \vspace{2ex}\\
\mathcal{L}_{total}^2 =\mathcal{L}_{d}^2 + \mathcal{L}_{t}^2
\end{array},
\right.
\end{equation}
where $L_{BCE}$ means the BCE loss.

In the third stage, we still use the PPA loss to compute the loss of the initial feature extraction subnet. In parallel, for the final high-quality prediction maps $\left\{\mathbf{P}_{f}^{i}\right\}_{i=0}^3$, we also adopt the PPA loss to compute the loss value, and for the edge prediction maps, we employ the BCE loss to compute the loss value of the ER module. The entire loss can be written as follows

\begin{equation}\label{eq-loss}
\left \{\begin{array}{l}
\mathcal{L}_{f}^3 =\sum_{i=0}^{3}{L_{PPA}(\mathbf{P}_{f}^{i}, \mathbf{GT})} \vspace{2ex}\\
\mathcal{L}_{eg}^3 =L_{BCE}(\mathbf{P}_{eg}, \mathbf{GT}_{eg}) \vspace{2ex}\\
\mathcal{L}_{total}^3 = \mathcal{L}_{total}^1 + \mathcal{L}_{f}^3 + \mathcal{L}_{eg}^3
\end{array},
\right.
\end{equation}
where $\mathbf{GT}_{eg}$ means the ground truth of edge.

\begin{table*}[htbp]
  \small
  \renewcommand{\arraystretch}{1.1}
  \renewcommand{\tabcolsep}{1.5mm}
    \centering
    \caption{Quantitative comparison results of $S$, ${F}_\beta^{max}$, ${F}_\beta^{mean}$, ${F}_\beta^{adp}$, ${E}_\xi^{max}$, ${E}_\xi^{mean}$, ${E}_\xi^{adp}$ and $MAE$ on the VDT-2048 dataset. Here, ``$\uparrow$" (``$\downarrow$") means that the larger (smaller) the better. The best three results in each row are marked in \textcolor[rgb]{ 1,  0,  0}{red}, \textcolor[rgb]{ 0,  .69,  .314}{green}, and \textcolor[rgb]{ 0,  .439,  .753}{blue}, respectively.}
    \begin{tabular}{p{5.69em}|c|cccccccc}
    \toprule
    Methods & Type  & $S\uparrow$     & $MAE\downarrow$   & ${E}_\xi^{adp}\uparrow$ & ${E}_\xi^{mean}\uparrow$ & ${E}_\xi^{max}\uparrow$ & ${F}_\beta^{adp}\uparrow$ & ${F}_\beta^{mean}\uparrow$ & ${F}_\beta^{max}\uparrow$ \\
    \midrule
    CPD   & V     & 0.9044  & 0.0039  & 0.9270  & 0.9501  & 0.9738  & 0.7645  & 0.8376  & 0.8603  \\
    RAS   & V     & 0.8900  & 0.0040  & 0.9615  & 0.9650  & 0.9688  & 0.8079  & 0.8292  & 0.8464  \\
    \midrule
    BBSNet & VD    & 0.9117  & 0.0046  & 0.8747  & 0.9357  & 0.9769  & 0.6957  & 0.8267  & 0.8679  \\
    DPANet & VD    & 0.7226  & 0.0192  & 0.5328  & 0.7222  & 0.8420  & 0.2919  & 0.4852  & 0.5733  \\
    RD3D  & VD    & 0.9095  & 0.0047  & 0.8354  & 0.9231  & 0.9766  & 0.6462  & 0.8120  & 0.8668  \\
    \midrule
    CGFNet & VT    & 0.9166  & 0.0033  & 0.9319  & 0.9447  & 0.9822  & 0.7822  & 0.8480  & 0.8817  \\
    CSRNet & VT    & 0.8821  & 0.0050  & 0.9494  & 0.9557  & 0.9656  & 0.7888  & 0.8278  & 0.8415  \\
    DCNet & VT    & 0.8787  & 0.0038  & 0.9658  & 0.9436  & 0.9659  & 0.8521  & 0.8450  & 0.8525  \\
    LSNet & VT    & 0.8867  & 0.0044  & 0.9327  & 0.9631  & 0.9737  & 0.7607  & 0.8097  & 0.8354  \\
    \midrule
    SwinNet & VD    & 0.9198  & 0.0037  & 0.8978  & 0.9507  & 0.9822  & 0.7321  & 0.8458  & 0.8796  \\
    SwinNet & VT    & \textcolor[rgb]{ 0,  .439,  .753}{0.9370} & \textcolor[rgb]{ 0,  .439,  .753}{0.0026} & 0.9444 & 0.9746 & \textcolor[rgb]{ 0,  .439,  .753}{0.9897} & 0.8090  & 0.8887 & 0.9095 \\
    \multicolumn{1}{l|}{HRTransNet} & VD    & 0.9144  & 0.0031  & 0.9617  & 0.9760  & 0.9844  & \textcolor[rgb]{ 0,  .69,  .314}{0.8827} & 0.8549  & 0.8802  \\
    \multicolumn{1}{l|}{HRTransNet} & VT    & 0.9281  & \textcolor[rgb]{ 0,  .439,  .753}{0.0026} & 0.9680  & 0.9809  & 0.9894  & 0.8446  & 0.8759  & 0.9008  \\
    \midrule
    HWSI  & VDT   & 0.9318  & \textcolor[rgb]{ 0,  .439,  .753}{0.0026} & \textcolor[rgb]{ 0,  .439,  .753}{0.9815} & \textcolor[rgb]{ 0,  .69,  .314}{0.9845} & 0.9890  & 0.8718  & \textcolor[rgb]{ 0,  .439,  .753}{0.8961} & \textcolor[rgb]{ 0,  .439,  .753}{0.9102} \\
    \multicolumn{1}{l|}{MFFNet} & VDT   & \textcolor[rgb]{ 0,  .69,  .314}{0.9394} & \textcolor[rgb]{ 0,  .69,  .314}{0.0025} & \textcolor[rgb]{ 0,  .69,  .314}{0.9831} & \textcolor[rgb]{ 0,  .439,  .753}{0.9825} & \textcolor[rgb]{ 0,  .69,  .314}{0.9905} & \textcolor[rgb]{ 0,  .439,  .753}{0.8758} & \textcolor[rgb]{ 0,  .69,  .314}{0.9034} & \textcolor[rgb]{ 0,  .69,  .314}{0.9180} \\
    \midrule
    \midrule
    Ours  & VDT   & \textcolor[rgb]{ 1,  0,  0}{0.9426} & \textcolor[rgb]{ 1,  0,  0}{0.0021} & \textcolor[rgb]{ 1,  0,  0}{0.9868} & \textcolor[rgb]{ 1,  0,  0}{0.9890} & \textcolor[rgb]{ 1,  0,  0}{0.9928} & \textcolor[rgb]{ 1,  0,  0}{0.8902} & \textcolor[rgb]{ 1,  0,  0}{0.9088} & \textcolor[rgb]{ 1,  0,  0}{0.9254} \\
    \bottomrule
    \end{tabular}%
  \label{tab_quantitative_comparison}%
\end{table*}%

\begin{table*}[htbp]
  \scriptsize
  \renewcommand{\arraystretch}{1.1}
  \renewcommand{\tabcolsep}{0.4mm}
    \centering
    \caption{Quantitative results in V challenges. Here, ``$\uparrow$" (``$\downarrow$") means that the larger (smaller) the better. The best three results in each row are marked in \textcolor[rgb]{ 1,  0,  0}{red}, \textcolor[rgb]{ 0,  .69,  .314}{green}, and \textcolor[rgb]{ 0,  .439,  .753}{blue}, respectively.}
    \resizebox{\linewidth}{!}{
    \begin{tabular}{p{6em}|ccc|ccc|ccc|ccc|ccc|ccc|ccc}
    \toprule
    \multicolumn{1}{l|}{\multirow{2}[4]{*}{Method}} & \multicolumn{3}{c|}{V-BSO} & \multicolumn{3}{c|}{V-LI} & \multicolumn{3}{c|}{V-MSO} & \multicolumn{3}{c|}{V-NI} & \multicolumn{3}{c|}{V-SA} & \multicolumn{3}{c|}{V-SI} & \multicolumn{3}{c}{V-SSO} \\
\cmidrule{2-22}    \multicolumn{1}{l|}{} & $MAE\downarrow$   & ${E}_\xi^{max}\uparrow$ & ${F}_\beta^{max}\uparrow$ & $MAE\downarrow$   & ${E}_\xi^{max}\uparrow$ & ${F}_\beta^{max}\uparrow$& $MAE\downarrow$   & ${E}_\xi^{max}\uparrow$ & ${F}_\beta^{max}\uparrow$& $MAE\downarrow$   & ${E}_\xi^{max}\uparrow$ & ${F}_\beta^{max}\uparrow$& $MAE\downarrow$   & ${E}_\xi^{max}\uparrow$ & ${F}_\beta^{max}\uparrow$& $MAE\downarrow$   & ${E}_\xi^{max}\uparrow$ & ${F}_\beta^{max}\uparrow$& $MAE\downarrow$   & ${E}_\xi^{max}\uparrow$ & ${F}_\beta^{max}\uparrow$ \\
    \midrule
    BBSNet & 0.0077  & 0.9909  & 0.9588  & 0.0064  & 0.9770  & 0.8386  & 0.0053  & 0.9814  & 0.8770  & 0.0080  & 0.9336  & 0.7179  & 0.0065  & 0.9753  & 0.8413  & 0.0065  & 0.9753  & 0.8413  & 0.0028  & 0.9645  & 0.7878  \\
    CGFNet & 0.0071  & 0.9930  & 0.9584  & 0.0046  & 0.9819  & 0.8628  & 0.0049  & 0.9764  & 0.8730  & 0.0035  & 0.9659  & 0.8223  & 0.0030  & 0.9866  & 0.8701  & 0.0043  & 0.9827  & 0.8735  & 0.0012  & 0.9591  & 0.7915  \\
    CPD   & 0.0092  & 0.9851  & 0.9456  & 0.0055  & 0.9750  & 0.8332  & 0.0046  & 0.9791  & 0.8705  & 0.0055  & 0.9300  & 0.6936  & 0.0030  & 0.9842  & 0.8826  & 0.0055  & 0.9701  & 0.8412  & 0.0012  & 0.9586  & 0.7760  \\
    CSRNet & 0.0139  & 0.9640  & 0.8999  & 0.0058  & 0.9700  & 0.8358  & 0.0089  & 0.9497  & 0.8271  & 0.0043  & 0.9525  & 0.7980  & 0.0047  & 0.9511  & 0.7887  & 0.0078  & 0.9489  & 0.8200  & 0.0014  & 0.9393  & 0.7377  \\
    DCNet & 0.0082  & 0.9907  & 0.9480  & 0.0048  & 0.9705  & 0.8430  & 0.0054  & 0.9800  & 0.8532  & 0.0038  & 0.9209  & 0.7834  & 0.0032  & 0.9833  & 0.8480  & 0.0047  & 0.9818  & 0.8613  & 0.0012  & 0.9266  & 0.7208  \\
    DPANet & 0.0418  & 0.9224  & 0.7191  & 0.0215  & 0.8317  & 0.5550  & 0.0264  & 0.8374  & 0.5780  & 0.0255  & 0.7233  & 0.3967  & 0.0166  & 0.8637  & 0.5692  & 0.0221  & 0.8336  & 0.5310  & 0.0108  & 0.8080  & 0.3958  \\
    LSNet & 0.0099  & 0.9901  & 0.9469  & 0.0057  & 0.9763  & 0.8253  & 0.0060  & 0.9778  & 0.8419  & 0.0052  & 0.9376  & 0.7044  & 0.0040  & 0.9747  & 0.8301  & 0.0064  & 0.9685  & 0.8036  & 0.0018  & 0.9533  & 0.6982  \\
    RAS   & 0.0095  & 0.9819  & 0.9377  & 0.0057  & 0.9739  & 0.8247  & 0.0055  & 0.9697  & 0.8432  & 0.0052  & 0.9156  & 0.6970  & 0.0033  & 0.9784  & 0.8593  & 0.0065  & 0.9615  & 0.8044  & 0.0013  & 0.9539  & 0.7510  \\
    RD3D  & 0.0091  & 0.9895  & 0.9552  & 0.0062  & 0.9775  & 0.8337  & 0.0057  & 0.9870  & 0.8764  & 0.0067  & 0.9383  & 0.7179  & 0.0038  & 0.9813  & 0.8829  & 0.0070  & 0.9702  & 0.8457  & 0.0024  & 0.9647  & 0.7785  \\
    SwinNet(VD) & 0.0068  & 0.9931  & 0.9634  & 0.0048  & 0.9782  & 0.8546  & 0.0048  & 0.9888  & 0.8809  & 0.0052  & 0.9541  & 0.7515  & 0.0028  & \textcolor[rgb]{ 0,  .69,  .314}{0.9932} & \textcolor[rgb]{ 0,  .439,  .753}{0.9041} & 0.0054  & 0.9736  & 0.8576  & 0.0024  & 0.9716  & 0.7995  \\
    SwinNet(VT) & \textcolor[rgb]{ 0,  .69,  .314}{0.0053} & \textcolor[rgb]{ 0,  .69,  .314}{0.9948} & \textcolor[rgb]{ 0,  .69,  .314}{0.9700} & \textcolor[rgb]{ 0,  .69,  .314}{0.0035} & 0.9884  & \textcolor[rgb]{ 0,  .69,  .314}{0.8976} & \textcolor[rgb]{ 0,  .439,  .753}{0.0040} & \textcolor[rgb]{ 0,  .69,  .314}{0.9919} & 0.9011  & \textcolor[rgb]{ 0,  .439,  .753}{0.0027} & 0.9774  & 0.8522  & \textcolor[rgb]{ 0,  .439,  .753}{0.0023} & \textcolor[rgb]{ 0,  .439,  .753}{0.9931} & 0.8991  & 0.0039  & 0.9868  & \textcolor[rgb]{ 0,  .439,  .753}{0.8948} & 0.0009  & 0.9805  & 0.8415  \\
    HRTrans(VD) & 0.0064  & 0.9927  & 0.9602  & 0.0042  & 0.9857  & 0.8659  & 0.0042  & 0.9892  & 0.8843  & 0.0042  & 0.9544  & 0.7533  & 0.0025  & 0.9922  & 0.9000  & 0.0049  & 0.9812  & 0.8635  & 0.0012  & 0.9823  & 0.7834  \\
    HRTrans(VT) & \textcolor[rgb]{ 0,  .439,  .753}{0.0057} & 0.9929  & 0.9643  & 0.0037  & \textcolor[rgb]{ 0,  .439,  .753}{0.9888} & 0.8873  & \textcolor[rgb]{ 0,  .69,  .314}{0.0039} & \textcolor[rgb]{ 0,  .439,  .753}{0.9916} & 0.8995  & 0.0030  & \textcolor[rgb]{ 0,  .439,  .753}{0.9781} & 0.8342  & \textcolor[rgb]{ 0,  .439,  .753}{0.0023} & \textcolor[rgb]{ 0,  .439,  .753}{0.9931} & \textcolor[rgb]{ 0,  .69,  .314}{0.9047} & 0.0042  & 0.9819  & 0.8777  & 0.0011  & \textcolor[rgb]{ 0,  .439,  .753}{0.9855} & 0.8083  \\
    HWSI  & 0.0061  & 0.9931  & 0.9618  & 0.0038  & 0.9859  & 0.8870  & \textcolor[rgb]{ 0,  .439,  .753}{0.0040} & 0.9885  & \textcolor[rgb]{ 0,  .439,  .753}{0.9055} & 0.0028  & 0.9754  & \textcolor[rgb]{ 0,  .439,  .753}{0.8584} & 0.0024  & 0.9860  & 0.8898  & \textcolor[rgb]{ 0,  .439,  .753}{0.0038} & \textcolor[rgb]{ 0,  .69,  .314}{0.9875} & 0.8916  & \textcolor[rgb]{ 0,  .439,  .753}{0.0008} & 0.9842  & \textcolor[rgb]{ 0,  .439,  .753}{0.8650} \\
    MFFNet & \textcolor[rgb]{ 0,  .439,  .753}{0.0057} & \textcolor[rgb]{ 0,  .439,  .753}{0.9937} & \textcolor[rgb]{ 0,  .439,  .753}{0.9656} & \textcolor[rgb]{ 0,  .69,  .314}{0.0035} & \textcolor[rgb]{ 0,  .69,  .314}{0.9889} & \textcolor[rgb]{ 0,  .439,  .753}{0.8975} & 0.0041  & 0.9906  & \textcolor[rgb]{ 0,  .69,  .314}{0.9147} & \textcolor[rgb]{ 0,  .69,  .314}{0.0025} & \textcolor[rgb]{ 0,  .69,  .314}{0.9796} & \textcolor[rgb]{ 0,  .69,  .314}{0.8695} & \textcolor[rgb]{ 0,  .69,  .314}{0.0022} & 0.9906  & 0.9022  & \textcolor[rgb]{ 0,  .69,  .314}{0.0036} & \textcolor[rgb]{ 0,  .439,  .753}{0.9873} & \textcolor[rgb]{ 0,  .69,  .314}{0.8970} & \textcolor[rgb]{ 1,  0,  0}{0.0007} & \textcolor[rgb]{ 1,  0,  0}{0.9886} & \textcolor[rgb]{ 0,  .69,  .314}{0.8739} \\
    \midrule
    \midrule
    Ours  & \textcolor[rgb]{ 1,  0,  0}{0.0044} & \textcolor[rgb]{ 1,  0,  0}{0.9956} & \textcolor[rgb]{ 1,  0,  0}{0.9730 } & \textcolor[rgb]{ 1,  0,  0}{0.0030} & \textcolor[rgb]{ 1,  0,  0}{0.9908} & \textcolor[rgb]{ 1,  0,  0}{0.9095} & \textcolor[rgb]{ 1,  0,  0}{0.0034} & \textcolor[rgb]{ 1,  0,  0}{0.9927} & \textcolor[rgb]{ 1,  0,  0}{0.9181} & \textcolor[rgb]{ 1,  0,  0}{0.0023} & \textcolor[rgb]{ 1,  0,  0}{0.9853} & \textcolor[rgb]{ 1,  0,  0}{0.8791} & \textcolor[rgb]{ 1,  0,  0}{0.0018} & \textcolor[rgb]{ 1,  0,  0}{0.9958} & \textcolor[rgb]{ 1,  0,  0}{0.9258} & \textcolor[rgb]{ 1,  0,  0}{0.0034} & \textcolor[rgb]{ 1,  0,  0}{0.9897} & \textcolor[rgb]{ 1,  0,  0}{0.9050} & \textcolor[rgb]{ 1,  0,  0}{0.0007} & \textcolor[rgb]{ 0,  .69,  .314}{0.9885} & \textcolor[rgb]{ 1,  0,  0}{0.8767} \\
    \bottomrule
    \end{tabular}%
    }
  \label{tab:com_v}%
\end{table*}%

\begin{table*}[htbp]
  \footnotesize
  \renewcommand{\arraystretch}{1.1}
  \renewcommand{\tabcolsep}{0.4mm}
    \centering
    \caption{Quantitative results in D challenges. Here, ``$\uparrow$" (``$\downarrow$") means that the larger (smaller) the better. The best three results in each row are marked in \textcolor[rgb]{ 1,  0,  0}{red}, \textcolor[rgb]{ 0,  .69,  .314}{green}, and \textcolor[rgb]{ 0,  .439,  .753}{blue}, respectively.}
    \begin{tabular}{p{6em}|cccc|cccc|cccc|cccc}
    \toprule
    \multicolumn{1}{l|}{\multirow{2}[4]{*}{Method}} & \multicolumn{4}{c|}{D-BI}     & \multicolumn{4}{c|}{D-BM}     & \multicolumn{4}{c|}{D-II}     & \multicolumn{4}{c}{D-SSO} \\
\cmidrule{2-17}    \multicolumn{1}{l|}{} & $S\uparrow$    & $MAE\downarrow$   & ${E}_\xi^{max}\uparrow$ & ${F}_\beta^{max}\uparrow$& $S\uparrow$    & $MAE\downarrow$   & ${E}_\xi^{max}\uparrow$ & ${F}_\beta^{max}\uparrow$& $S\uparrow$    & $MAE\downarrow$   & ${E}_\xi^{max}\uparrow$ & ${F}_\beta^{max}\uparrow$& $S\uparrow$    & $MAE\downarrow$   & ${E}_\xi^{max}\uparrow$ & ${F}_\beta^{max}\uparrow$ \\
    \midrule
    BBSNet & 0.9067  & 0.0046  & 0.9747  & 0.8588  & 0.9027  & 0.0041  & 0.9766  & 0.8565  & 0.9274  & 0.0046  & 0.9829  & 0.8966  & 0.8548  & 0.0028  & 0.9645  & 0.7878  \\
    DPANet & 0.7068  & 0.0189  & 0.8333  & 0.5521  & 0.7226  & 0.0182  & 0.8395  & 0.5480  & 0.7665  & 0.0206  & 0.8622  & 0.6342  & 0.6204  & 0.0108  & 0.8080  & 0.3958  \\
    RD3D  & 0.9031  & 0.0046  & 0.9738  & 0.8565  & 0.9076  & 0.0044  & 0.9799  & 0.8626  & 0.9289  & 0.0052  & 0.9845  & 0.8988  & 0.8466  & 0.0024  & 0.9647  & 0.7785  \\
    SwinNet(VD) & 0.9155  & 0.0034  & 0.9818  & 0.8726  & 0.9127  & 0.0044  & 0.9782  & 0.8660  & 0.9330  & 0.0047  & 0.9842  & 0.9023  & 0.8613  & 0.0024  & 0.9716  & 0.7995  \\
    HRTrans(VD) & 0.9092  & 0.0029  & 0.9849  & 0.8723  & 0.9081  & 0.0032  & 0.9758  & 0.8650  & 0.9303  & 0.0036  & 0.9832  & 0.9043  & 0.8463  & \textcolor[rgb]{ 0,  .439,  .753}{0.0012} & 0.9823  & 0.7834  \\
    HWSI  & \textcolor[rgb]{ 0,  .439,  .753}{0.9286} & \textcolor[rgb]{ 0,  .69,  .314}{0.0023} & \textcolor[rgb]{ 0,  .439,  .753}{0.9882} & \textcolor[rgb]{ 0,  .439,  .753}{0.9057} & \textcolor[rgb]{ 0,  .439,  .753}{0.9285} & \textcolor[rgb]{ 0,  .439,  .753}{0.0029} & \textcolor[rgb]{ 0,  .439,  .753}{0.9889} & \textcolor[rgb]{ 0,  .439,  .753}{0.9020} & \textcolor[rgb]{ 0,  .439,  .753}{0.9417} & \textcolor[rgb]{ 0,  .439,  .753}{0.0035} & \textcolor[rgb]{ 0,  .439,  .753}{0.9909} & \textcolor[rgb]{ 0,  .439,  .753}{0.9249} & \textcolor[rgb]{ 1,  0,  0}{0.9253} & 0.0025  & \textcolor[rgb]{ 0,  .439,  .753}{0.9848} & \textcolor[rgb]{ 1,  0,  0}{0.8972} \\
    MFFNet & \textcolor[rgb]{ 0,  .69,  .314}{0.9364} & \textcolor[rgb]{ 0,  .69,  .314}{0.0023} & \textcolor[rgb]{ 0,  .69,  .314}{0.9900} & \textcolor[rgb]{ 0,  .69,  .314}{0.9132} & \textcolor[rgb]{ 0,  .69,  .314}{0.9327} & \textcolor[rgb]{ 0,  .69,  .314}{0.0026} & \textcolor[rgb]{ 0,  .69,  .314}{0.9901} & \textcolor[rgb]{ 0,  .69,  .314}{0.9096} & \textcolor[rgb]{ 0,  .69,  .314}{0.9461} & \textcolor[rgb]{ 0,  .69,  .314}{0.0032} & \textcolor[rgb]{ 0,  .69,  .314}{0.9919} & \textcolor[rgb]{ 0,  .69,  .314}{0.9311} & \textcolor[rgb]{ 0,  .69,  .314}{0.9047} & \textcolor[rgb]{ 1,  0,  0}{0.0007} & \textcolor[rgb]{ 1,  0,  0}{0.9886} & \textcolor[rgb]{ 0,  .439,  .753}{0.8739} \\
    \midrule
    \midrule
    Ours  & \textcolor[rgb]{ 1,  0,  0}{0.9395} & \textcolor[rgb]{ 1,  0,  0}{0.0020} & \textcolor[rgb]{ 1,  0,  0}{0.9925} & \textcolor[rgb]{ 1,  0,  0}{0.9207} & \textcolor[rgb]{ 1,  0,  0}{0.9392} & \textcolor[rgb]{ 1,  0,  0}{0.0022} & \textcolor[rgb]{ 1,  0,  0}{0.9918} & \textcolor[rgb]{ 1,  0,  0}{0.9201} & \textcolor[rgb]{ 1,  0,  0}{0.9527} & \textcolor[rgb]{ 1,  0,  0}{0.0025} & \textcolor[rgb]{ 1,  0,  0}{0.9938} & \textcolor[rgb]{ 1,  0,  0}{0.9407} & \textcolor[rgb]{ 0,  .439,  .753}{0.8948} & \textcolor[rgb]{ 1,  0,  0}{0.0007} & \textcolor[rgb]{ 0,  .69,  .314}{0.9885} & \textcolor[rgb]{ 0,  .69,  .314}{0.8767} \\
    \bottomrule
    \end{tabular}%
  \label{tab:com_d}%
\end{table*}%

\begin{table*}[htbp]
  \footnotesize
  \renewcommand{\arraystretch}{1.1}
  \renewcommand{\tabcolsep}{0.4mm}
    \centering
    \caption{Quantitative results in T challenges. Here, ``$\uparrow$" (``$\downarrow$") means that the larger (smaller) the better. The best three results in each row are marked in \textcolor[rgb]{ 1,  0,  0}{red}, \textcolor[rgb]{ 0,  .69,  .314}{green}, and \textcolor[rgb]{ 0,  .439,  .753}{blue}, respectively.}
    \begin{tabular}{p{6em}|cccc|cccc|cccc}
    \toprule
    \multicolumn{1}{l|}{\multirow{2}[4]{*}{Method}} & \multicolumn{4}{c|}{T-Cr}     & \multicolumn{4}{c|}{T-HR}     & \multicolumn{4}{c}{T-RD} \\
\cmidrule{2-13}    \multicolumn{1}{l|}{} & $S\uparrow$    & $MAE\downarrow$   & ${E}_\xi^{max}\uparrow$ & ${F}_\beta^{max}\uparrow$& $S\uparrow$    & $MAE\downarrow$   & ${E}_\xi^{max}\uparrow$ & ${F}_\beta^{max}\uparrow$& $S\uparrow$    & $MAE\downarrow$   & ${E}_\xi^{max}\uparrow$ & ${F}_\beta^{max}\uparrow$\\
    \midrule
    CGFNet & 0.9010  & 0.0034  & 0.9738  & 0.8590  & 0.9523  & 0.0029  & 0.9921  & 0.9259  & 0.9302  & 0.0046  & 0.9902  & 0.9081  \\
    CSRNet & 0.8453  & 0.0050  & 0.9499  & 0.7955  & 0.9374  & 0.0033  & 0.9919  & 0.9118  & 0.8981  & 0.0061  & 0.9812  & 0.8726  \\
    DCNet & 0.8582  & 0.0039  & 0.9511  & 0.8215  & 0.9305  & 0.0031  & 0.9895  & 0.9072  & 0.9038  & 0.0051  & 0.9835  & 0.8904  \\
    LSNet & 0.8808  & 0.0042  & 0.9689  & 0.8297  & 0.9095  & 0.0043  & 0.9788  & 0.8740  & 0.9011  & 0.0064  & 0.9765  & 0.8582  \\
    SwinNet(VT) & \textcolor[rgb]{ 0,  .439,  .753}{0.9263} & 0.0026  & 0.9852  & 0.8956  & \textcolor[rgb]{ 0,  .439,  .753}{0.9608} & \textcolor[rgb]{ 0,  .439,  .753}{0.0021} & 0.9950  & \textcolor[rgb]{ 0,  .439,  .753}{0.9428} & \textcolor[rgb]{ 1,  0,  0}{0.9434} & \textcolor[rgb]{ 0,  .69,  .314}{0.0037} & \textcolor[rgb]{ 0,  .69,  .314}{0.9922} & \textcolor[rgb]{ 0,  .439,  .753}{0.9220} \\
    HRTrans(VT) & 0.9130  & 0.0027  & \textcolor[rgb]{ 0,  .69,  .314}{0.9875} & 0.8856  & 0.9482  & 0.0022  & \textcolor[rgb]{ 1,  0,  0}{0.9955} & 0.9312  & 0.9392  & \textcolor[rgb]{ 0,  .439,  .753}{0.0038} & \textcolor[rgb]{ 0,  .69,  .314}{0.9922} & 0.9173  \\
    HWSI  & 0.9253  & \textcolor[rgb]{ 0,  .439,  .753}{0.0025} & 0.9848  & \textcolor[rgb]{ 0,  .439,  .753}{0.8972} & 0.9470  & 0.0026  & 0.9941  & 0.9320  & 0.9324  & 0.0040  & \textcolor[rgb]{ 1,  0,  0}{0.9925} & 0.9152  \\
    MFFNet & \textcolor[rgb]{ 0,  .69,  .314}{0.9312} & \textcolor[rgb]{ 0,  .69,  .314}{0.0023} & \textcolor[rgb]{ 0,  .439,  .753}{0.9866} & \textcolor[rgb]{ 0,  .69,  .314}{0.9052} & \textcolor[rgb]{ 1,  0,  0}{0.9610} & \textcolor[rgb]{ 0,  .69,  .314}{0.0019} & \textcolor[rgb]{ 0,  .69,  .314}{0.9954} & \textcolor[rgb]{ 0,  .69,  .314}{0.9472} & \textcolor[rgb]{ 0,  .439,  .753}{0.9394} & \textcolor[rgb]{ 0,  .439,  .753}{0.0038} & 0.9912  & \textcolor[rgb]{ 0,  .69,  .314}{0.9233} \\
    \midrule
    \midrule
    Ours  & \textcolor[rgb]{ 1,  0,  0}{0.9341} & \textcolor[rgb]{ 1,  0,  0}{0.0021} & \textcolor[rgb]{ 1,  0,  0}{0.9918} & \textcolor[rgb]{ 1,  0,  0}{0.9159} & \textcolor[rgb]{ 0,  .69,  .314}{0.9609} & \textcolor[rgb]{ 1,  0,  0}{0.0018} & \textcolor[rgb]{ 0,  .439,  .753}{0.9952} & \textcolor[rgb]{ 1,  0,  0}{0.9500} & \textcolor[rgb]{ 0,  .69,  .314}{0.9406} & \textcolor[rgb]{ 1,  0,  0}{0.0034} & 0.9918  & \textcolor[rgb]{ 1,  0,  0}{0.9258} \\
    \bottomrule
    \end{tabular}%
  \label{tab:com_t}%
\end{table*}%

\section{Experimental Results}
\label{sec:experiment}
In this part, we first present the public VDT-2048 dataset and the implementation details in Section~\ref{Datasets and Implementation}. Secondly, the evaluation metrics are stated in Section~\ref{Evaluation Metrics}. Thirdly, in Section~\ref{Comparison with the State-of-the-arts}, some comparisons are conducted between the proposed QSF-Net and the state-of-the-art saliency models. Then, the ablation study is presented in Section~\ref{Ablation Studies}.  {Lastly, we discuss the failure cases and analysis in Section~\ref{sec:Failure}.}

\subsection{Datasets and Implementation Details}
\label{Datasets and Implementation}
To comprehensively validate our model, we conduct extensive comparisons on VDT-2048 dataset\cite{song2022novel}. It contains 2048 images and the corresponding pixel-wise annotations (\emph{i.e.}, ground truths), of which 1048 images are used for training and 1000 images are used for testing.

We implement our model by using PyTorch\cite{paszke2019pytorch} on a PC equipped with an Intel(R) Core(TM) i9-12900K CPU and an RTX 3090 GPU. During the training phase, each image is resized to 384$\times$384, and data augmentation strategies are adopted including random flipping, rotation, and border clipping. The encoders of the initial feature extraction subnet and the quality-aware region selection subnet of our model are initialized by Swin-B\cite{liu2021swin} and ResNet-34\cite{he2016deep}, respectively. Other parameters are initialized randomly. Here, we employ Adam \cite{kingma2014adam} to optimize our network, where the batch size and the initial learning rate are set to 4 and $1\times 10^{-4}$, respectively.

%
\begin{figure*}[htbp]
\centering
\includegraphics[width=0.98\textwidth]{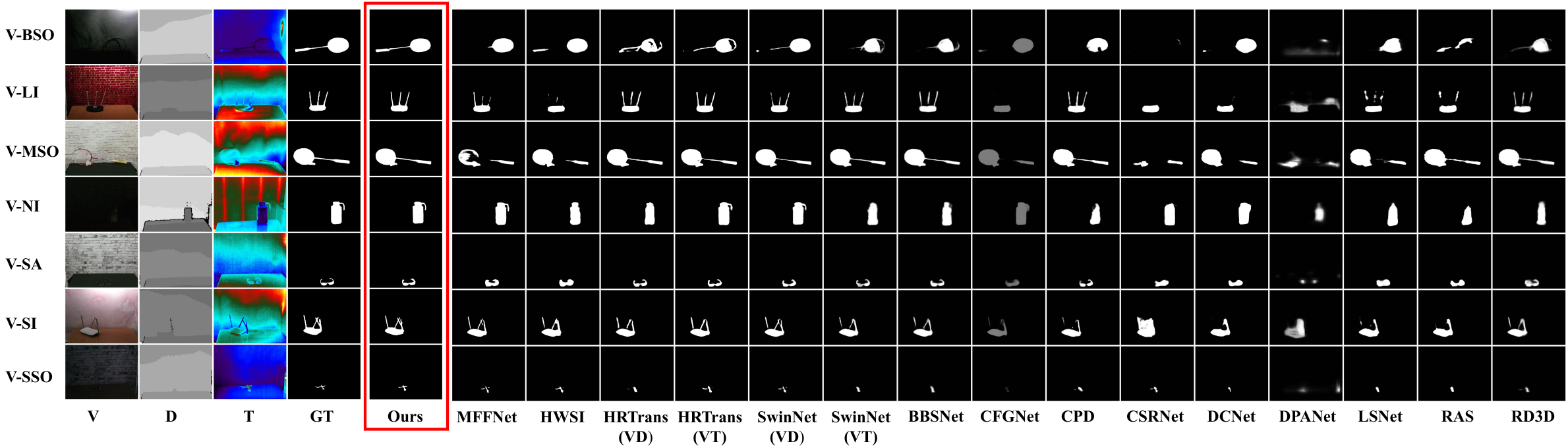}
\caption{ {Visual comparison of V-challenge.}} 
\label{fig_qualitative_v}
\end{figure*}
\begin{figure*}[htbp]
\centering
\includegraphics[width=0.98\textwidth]{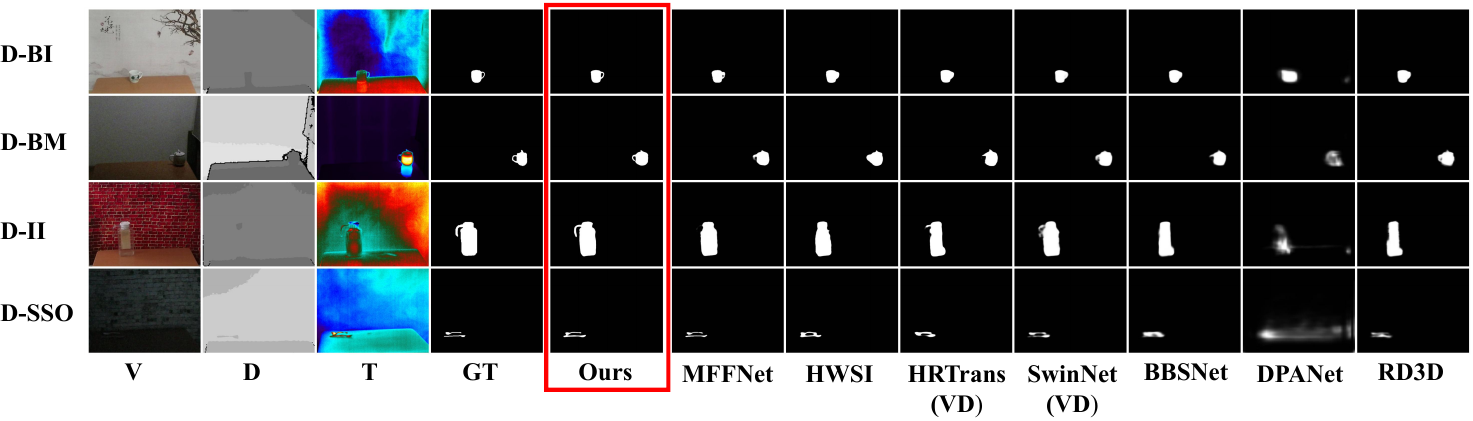}
\caption{ {Visual comparison of D-challenge.}} 
\label{fig_qualitative_d}
\end{figure*}
\begin{figure*}[htbp]
\centering
\includegraphics[width=0.98\textwidth]{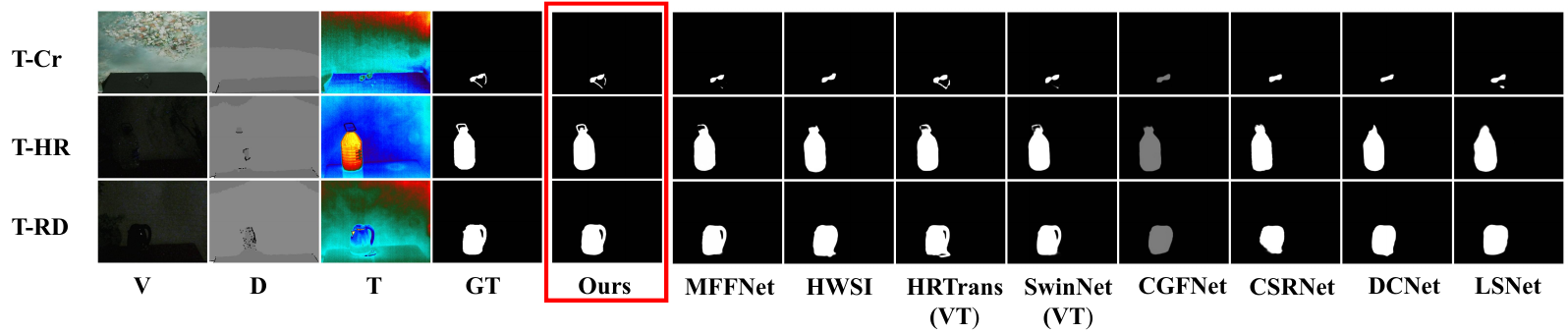}
\caption{ {Visual comparison of T-challenge.}} 
\label{fig_qualitative_t}
\end{figure*}
\begin{figure}[!t]
  \centering
\begin{tabular}{c@{\hspace{10pt}}c}
     \includegraphics[width = 0.22\textwidth]{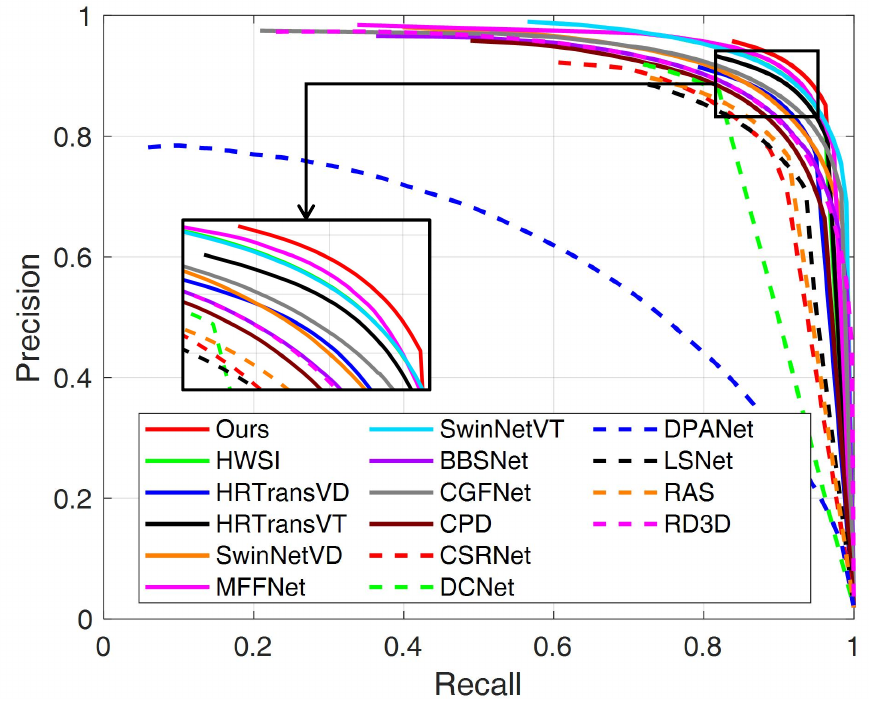}&
     \includegraphics[width = 0.22\textwidth]{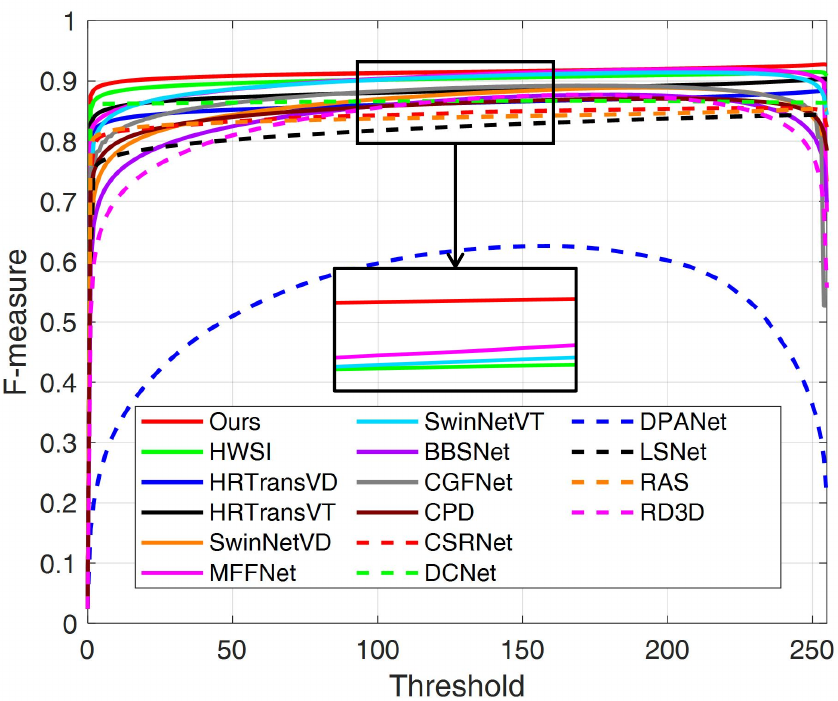} 
\end{tabular}
\caption{\small{(better viewed in color) Quantitative evaluation of different saliency models: the left part presents PR curves and the right part presents F-measure curves on VDT-2048 dataset.}}
\label{fig_Quantitative}
\end{figure}

\begin{table}[htbp]
  \centering
  \caption{ {Computational complexity comparison results of FPS, Parameters, and FLOPs.Here, ``$\uparrow$" (``$\downarrow$") means that the larger (smaller) the better.}}
    \begin{tabular}{p{4.875em}|c|c|c|c}
    \toprule
    \multicolumn{1}{r|}{} & Type  & Param(M)$\downarrow$ & FLOPs(G)$\downarrow$ & FPS$\uparrow$ \\
    \midrule
    RD3D  & VD    & 28.92  & 43.30  & 60.66  \\
    BBSNet & VD    & 49.77  & 31.32  & 46.26  \\
    DPANet & VD    & 92.39  & 58.96  & 56.63  \\
    \midrule
    CGFNet & VT    & 69.92  & 347.78  & 19.47  \\
    CSRNet & VT    & 1.01  & 4.36  & 51.54  \\
    DCNet & VT    & 24.06  & 207.21  & 37.67  \\
    LSNet & VT    & 4.57  & 1.23  & 123.50  \\
    \midrule
    SwinNet & VD/VT & 124.72  & 198.78  & 29.21  \\
    \multicolumn{1}{l|}{HRT} & VD/VT & 17.27  & 58.89  & 16.47  \\
    \midrule
    HWSI  & VDT   & 100.77  & 357.93  & 4.33  \\
    \multicolumn{1}{l|}{MFFNet} & VDT   & 103.24  & 1189.12  & 5.94  \\
    \midrule
    \midrule
    \textbf{Ours} & \textbf{VDT} & \textbf{159.72} & \textbf{280.60} & \textbf{11.16} \\
    \bottomrule
    \end{tabular}%
  \label{tab:cc}%
\end{table}%

\subsection{Evaluation Metrics}
\label{Evaluation Metrics}
To quantitatively make a comparison for different saliency models on VDT-2048 dataset, We evaluate our model using the 10 most extensive evaluation metrics, including S-measure ($S$)\cite{fan2017structure}, Mean Absolute Error ($MAE$)\cite{perazzi2012saliency}, F-measure (${F}_\beta^{max}$, ${F}_\beta^{mean}$, ${F}_\beta^{adp}$)\cite{achanta2009frequency}, E-measure (${E}_\xi^{max}$, ${E}_\xi^{mean}$, ${E}_\xi^{adp}$)\cite{fan2018enhanced}, F-measure curve, and Precision-Recall (PR) curve\cite{achanta2009frequency}.



\subsection{Comparison with the State-of-the-art Methods}
\label{Comparison with the State-of-the-arts}
To verify the effectiveness of our model, we compare our model with 13 state-of-the-art SOD models, which can be divided into five categories. Concretely, the first one contains two RGB SOD methods including CPD\cite{wu2019cascaded} and RAS\cite{chen2018reverse}. The second one consists of three RGB-D SOD methods including BBSNet\cite{fan2020bbs}, DPANet\cite{chen2020dpanet}, and RD3D\cite{chen20223}. The third one contains four RGB-T SOD methods including CGFNet\cite{wang2021cgfnet}, CSRNet\cite{huo2021efficient}, DCNet\cite{tu2022weakly}, and LSNet\cite{zhou2023lsnet}. The fourth one contains two transformer-based methods targeting both RGB-D and RGB-T SOD task SwinNet\cite{liu2021swinnet} and HRTransNet\cite{tang2022hrtransnet}. The fifth one contains two VDT SOD methods, $\emph{i.e.}$, HWSI\cite{song2022novel} and MFFNet\cite{wan2023mffnet}. For a fair comparison, the prediction results of all models are provided by the authors or are generated by running the source codes with their default settings.


\subsubsection{{\bf Quantitative comparison}}
To quantitatively evaluate our model on VDT-2048 datasets, we compare our model and other models in terms of $S$, ${F}_\beta^{max}$, ${F}_\beta^{mean}$, ${F}_\beta^{adp}$, ${E}_\xi^{max}$, ${E}_\xi^{mean}$, ${E}_\xi^{adp}$ and $MAE$, as presented in Tab.~\ref{tab_quantitative_comparison}. 
It can be found that our model performs better than all models in terms of all evaluation metrics. Particularly, compared with the second-best model MFFNet, our model presents better performance, where the performance is improved by 16.00\% in terms of $MAE$, 1.64\% in terms of ${F}_\beta^{adp}$, and 0.66\% in terms of ${E}_\xi^{mean}$. Meanwhile, we also provided the PR curves and F-measure curves of our model and several top-level multi-modal SOD models, as presented in Fig.~\ref{fig_Quantitative}. We can observe that the PR curve of our model is the closest one to the upper right corner, and the area below the F-measure curve of our model is also the largest one. From the above experiments, we can prove the effectiveness and superiority of our model.

Furthermore, to illustrate the robustness of our model in dealing with challenging scenarios, we conduct extensive experiments in various complex scenarios, which contain V-challenge ($\emph{i.e.}$, big salient object (BSO), low illumination (LI), multiple salient object (MSO), no illumination (NI), similar appearance (SA), side illumination (SI), and small salient object (SSO)), D-challenge ($\emph{i.e.}$, background interference (BI), background messy (BM), information incomplete (II), and small salient object (SSO)), and T-challenge ($\emph{i.e.}$, crossover (Cr), heat reflection (HR), and radiation dispersion (RD)). The results of V-challenge, D-challenge, and T challenge are shown in Table~\ref{tab:com_v}, Table~\ref{tab:com_d}, and Table~\ref{tab:com_t}, respectively. From the above results, it can be seen that our model achieves the best performance in most of the challenge scenarios. This indicates that our model can well deal with the challenging scenes, where one modality of the scene is low-quality. The reason behind this lies in the design of quality perception. We can find that quality-aware region selection subnet can acquire the valuable information of different modality features (V-D and V-T), which guides the fusion of different modality features.

 {In addition, to evaluate the computational efficiency, we make a comparison between our model and the state-of-the-art models in terms of parameters, FLOPs, and FPS, which are measured in million (M), giga (G), and frames per second (FPS), as presented in Table.~\ref{tab:cc}. Here, ``parameters'' denotes the total parameters of the network, which is a pivotal indicator of storage resource consumption. ``FLOPs” refers to the floating-point operations, serving as a quantitative metric to gauge the computational complexity of the network.
``Speed” represents the number of frames processed by the network per second, offering a direct reflection of the network's inference speed.
It can be found that though our model is with largest parameters, our model achieves comparable performance in terms of FLOPs and FPS, where the two metrics are at the intermediate level. Besides, compared other two VDT SOD models including HWSI and MFFNet, our model presents the smallest FLOPs and largest FPS. From the comparison results, we can say that there is still a large room to elevate the computational efficiency of our model.
}

\subsubsection{{\bf Qualitative comparison}}
To qualitatively make a comparison for all saliency models, some visual results of challenging scenes are presented in Fig.~\ref{fig_qualitative_v}, Fig.~\ref{fig_qualitative_d}, and Fig.~\ref{fig_qualitative_t}. It can be found that the prediction results of our model are more complete and accurate than other models. Specifically, in Fig.~\ref{fig_qualitative_v}, namely the V-challenge, due to poor illumination of RGB images, we can find that most SOD models fail to detect the complete salient objects, as shown in $1^{st}$, $2^{nd}$, $4^{th}$, $7^{th}$ rows in Fig.~\ref{fig_qualitative_v}. Besides, in Fig.~\ref{fig_qualitative_d}, namely the D-challenge, the performance of many models degrades largely in boundary details, because of the low differentiation between background and foreground in depth images, as presented in $2^{nd}$, $3^{rd}$, $4^{th}$ rows. Furthermore, influenced by thermal crossover, heat reflection, and radiation dispersion in thermal images, a portion of the salient objects whose temperature is close to the background is ignored, as shown in $1^{st}$, $2^{nd}$, $3^{rd}$ rows of Fig.~\ref{fig_qualitative_t}.

In our model, we utilize a quality-aware strategy to select regions in the depth and thermal images, which provides high-quality regions and low-quality regions to purify the initial modality features. After that, the two modalities can provide effective complementary information for the fusion of different modality features, which tackles the challenging situation when one of the modalities is in low-quality. In addition, in the last part, namely the region-guided selective fusion subnet, we explore the intra-modal and inter-modal correlations between different features via the IIA module, and endow the features with more spatial details via the ER module.  {This ensures the complete and  accurate detection of salient objects to some degree.} In conclusion, according to the quantitative and qualitative comparisons, we are able to unequivocally prove the superiority and efficiency of the proposed QSF-Net, where our model can highlight complete salient objects with accurate boundary details from VDT images.

\begin{figure*}[!t]
  \centering
\begin{tabular}{c@{\hspace{10pt}}c@{\hspace{10pt}}c@{\hspace{10pt}}c}
     \includegraphics[width = 0.23\textwidth]{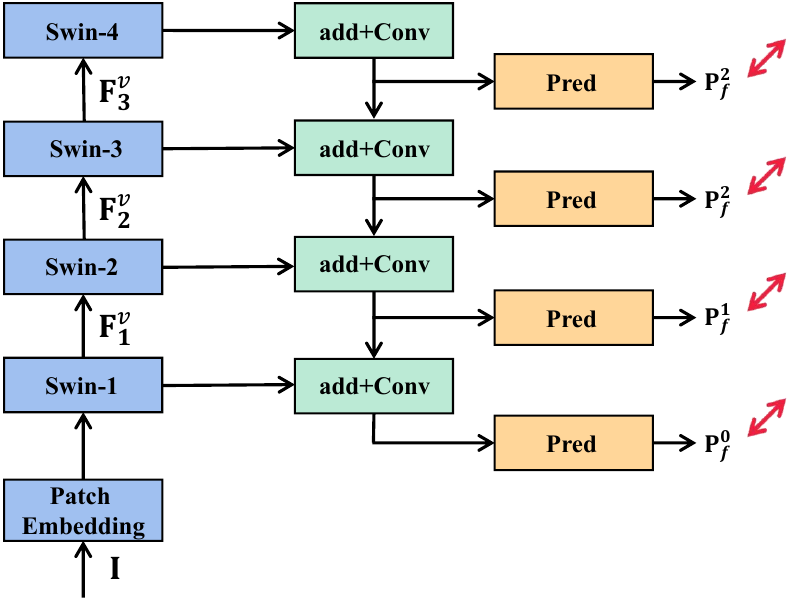}&
     \includegraphics[width = 0.23\textwidth]{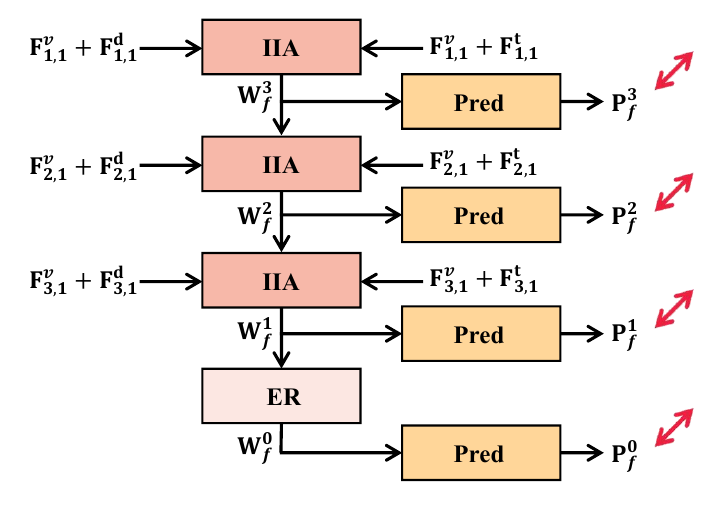}&
     \includegraphics[width = 0.23\textwidth]{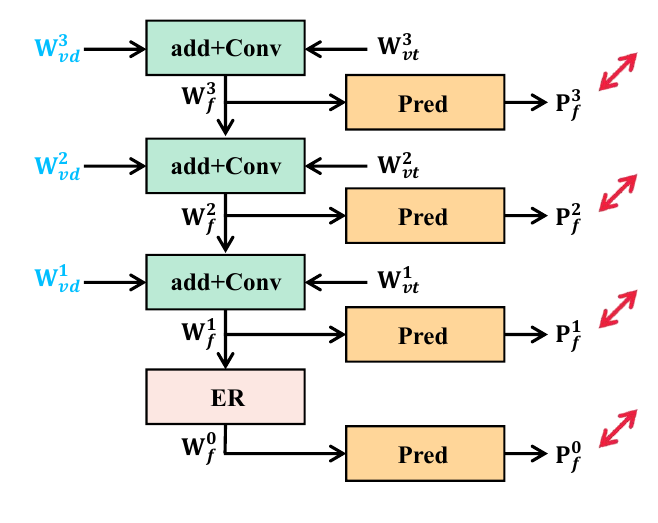}&
     \includegraphics[width = 0.23\textwidth]{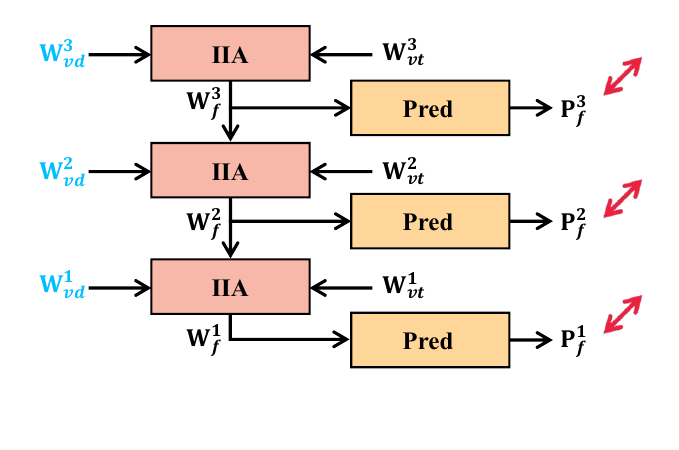}\\
       {(a) Base} &  {(b) w/o QA} &  {(c) w/o IIA} &  {(d) w/o ER} \\
\end{tabular}
\caption{The architecture of the variants in ablation studies.}
\label{fig_ab}
\end{figure*}

\subsection{Ablation Studies}
\label{Ablation Studies}

In this section, we will conduct experiments to validate the effectiveness of the key components of our model on the VDT-2048 dataset. 

Specifically, firstly, in the initial feature extraction subnet, to validate the necessity of our feature shrinkage pyramid structure with MSF module, we design six variants including ``Base\_V", ``Base\_D", ``Base\_T", ``Ours\_V", ``Ours\_D", and ``Ours\_T". In the former three variants, namely ``Base\_V", ``Base\_D", and ``Base\_T", we designed a U-shape network as a baseline, where the encoder is Swin Transformer and the decoder is convolution layers, as shown in Fig~\ref{fig_ab} (a). The difference among them is only the input images, which can be the RGB image (``Base\_V"), the depth image (``Base\_D"), or the thermal image (``Base\_T"). For the latter three variants including ``Ours\_V", ``Ours\_D", and ``Ours\_T", we utilize the single branch as the initial feature extraction subnet, whose input image is the RGB image (``Ours\_V"), the depth image (``Ours\_D"), or the thermal image (``Ours\_T"), respectively. 

 {Secondly, to validate the effectiveness of our quality-aware region selection subnet, we design three variants, \emph{i.e.,} ``w/o QA", ``w/o LQ", and ``w/o HQ". As shown in Fig~\ref{fig_ab} (b), ``w/o QA" denotes that we remove the quality-aware region selection subnet, and directly fuse the features $\left\{\mathbf{F}_{i,1}^{v}\right\}_{i=1}^3$, $\left\{\mathbf{F}_{i,1}^{d}\right\}_{i=1}^3$, and $\left\{\mathbf{F}_{i,1}^{t}\right\}_{i=1}^3$ via the element-wise summation, where the fused results are then sent to the IIA modules and ER module. ``w/o LQ" means that we train the quality-aware region selection subnet by using high-quality region $\mathbf{PGT}_{1}$ (Eq.~\ref{eq3}) based pseudo-GT only. Besides, to remove the impact of low-quality regions on the fusion process (Eq.~\ref{eq7}), we replace the $\mathbf{W}_{vd}^{i,2}$ with $\mathbf{F}_{i,1}^{v}$. Similarly, ``w/o HQ" refers to that we train the quality-aware region selection subnet by using low-quality region $\mathbf{PGT}_{2}$ (Eq.~\ref{eq4}) based pseudo-GT only. To remove the impact of high-quality regions on the fusion process (Eq.~\ref{eq6}), we replace the $\mathbf{W}_{vd}^{i,1}$ with $\mathbf{F}_{i,1}^{d}$.}

\begin{table}[!t]
  \centering
  \footnotesize
  \renewcommand{\arraystretch}{1}
  \renewcommand{\tabcolsep}{0.6mm}
  \caption{ {Ablation studies of initial feature extraction subnet. Here, ``$\uparrow$" (``$\downarrow$") means that the larger (smaller) the better.}}
  \resizebox{\linewidth}{!}{
    \begin{tabular}{l|cccccccc}
    \toprule
          & $S\uparrow$     & $MAE\downarrow$   & ${E}_\xi^{adp}\uparrow$ & ${E}_\xi^{mean}\uparrow$ & ${E}_\xi^{max}\uparrow$ & ${F}_\beta^{adp}\uparrow$ & ${F}_\beta^{mean}\uparrow$ & ${F}_\beta^{max}\uparrow$ \\
    \midrule
    Base\_V & 0.9106  & 0.0035  & 0.9517  & 0.9704  & 0.9782  & 0.8062  & 0.8471  & 0.8724  \\
    Base\_D & 0.7194  & 0.0128  & 0.7615  & 0.8309  & 0.8779  & 0.4952  & 0.5287  & 0.5489  \\
    Base\_T & 0.9055  & 0.0039  & 0.9539  & 0.9731  & 0.9818  & 0.7960  & 0.8351  & 0.8615  \\
    Ours\_V & 0.9176  & 0.0030  & 0.9744  & 0.9728  & 0.9785  & 0.8518  & 0.8717  & 0.8861  \\
    Ours\_D & 0.7074  & 0.0108  & 0.8497  & 0.8661  & 0.8733  & 0.5114  & 0.5217  & 0.5286  \\
    Ours\_T & 0.9119  & 0.0034  & 0.9778  & 0.9786  & 0.9825  & 0.8469  & 0.8627  & 0.8759  \\
    \bottomrule
    \end{tabular}%
    }
  \label{tab:abl}%
\end{table}%

\begin{table}[!t]
  \centering
  \footnotesize
  \renewcommand{\arraystretch}{1}
  \renewcommand{\tabcolsep}{0.6mm}
  \caption{ {Ablation studies of quality-aware region selection subnet. Here, ``$\uparrow$" (``$\downarrow$") means that the larger (smaller) the better.}}
  \resizebox{\linewidth}{!}{
    \begin{tabular}{p{4.19em}|cccccccc}
    \toprule
    & $S\uparrow$     & $MAE\downarrow$   & ${E}_\xi^{adp}\uparrow$ & ${E}_\xi^{mean}\uparrow$ & ${E}_\xi^{max}\uparrow$ & ${F}_\beta^{adp}\uparrow$ & ${F}_\beta^{mean}\uparrow$ & ${F}_\beta^{max}\uparrow$\\
    \midrule
    w/o QA & 0.9379  & 0.0023  & 0.9761  & 0.9839  & 0.9894  & 0.8640  & 0.8930  & 0.9178  \\
    w/o LQ & 0.9410  & 0.0022  & 0.9858  & 0.9882  & 0.9921  & 0.8868  & 0.9071  & 0.9234  \\
    w/o HQ & 0.9392  & 0.0023  & 0.9802  & 0.9864  & 0.9914  & 0.8729  & 0.8971  & 0.9177  \\
    \midrule
    \midrule
    \textbf{Ours} & \textbf{0.9426} & \textbf{0.0021} & \textbf{0.9868} & \textbf{0.9890} & \textbf{0.9928} & \textbf{0.8902} & \textbf{0.9088} & \textbf{0.9254} \\
    \bottomrule
    \end{tabular}%
    }
  \label{tab:ab_QRS}%
\end{table}%

%
\begin{table}[!t]
  \centering
  \footnotesize
  \renewcommand{\arraystretch}{1}
  \renewcommand{\tabcolsep}{0.6mm}
  \caption{ {Ablation studies of region-guided selective fusion subnet. Here, ``$\uparrow$" (``$\downarrow$") means that the larger (smaller) the better.}}
  \resizebox{\linewidth}{!}{
    \begin{tabular}{p{4.19em}|cccccccc}
    \toprule
    & $S\uparrow$     & $MAE\downarrow$   & ${E}_\xi^{adp}\uparrow$ & ${E}_\xi^{mean}\uparrow$ & ${E}_\xi^{max}\uparrow$ & ${F}_\beta^{adp}\uparrow$ & ${F}_\beta^{mean}\uparrow$ & ${F}_\beta^{max}\uparrow$ \\
    \midrule
    w/o IIA & 0.9391  & 0.0023  & 0.9806  & 0.9873  & 0.9921  & 0.8728  & 0.8990  & 0.9178  \\
    w/o ER & 0.9393  & 0.0023  & 0.9808  & 0.9872  & 0.9920  & 0.8742  & 0.8991  & 0.9173  \\
    \midrule
    \midrule
    \textbf{Ours} & \textbf{0.9426} & \textbf{0.0021} & \textbf{0.9868} & \textbf{0.9890} & \textbf{0.9928} & \textbf{0.8902} & \textbf{0.9088} & \textbf{0.9254} \\
    \bottomrule
    \end{tabular}%
    }
  \label{tab:ab2}%
\end{table}%

Lastly, to evaluate the effectiveness of two components in region-guided selective fusion subnet, \emph{i.e.}, IIA module, and ER module, we design two variants. The first variant is our region-guided selective fusion subnet without IIA module, namely ``w/o IIA", where we replace the IIA module with element-wise summation and convolutional layers, as shown in Fig~\ref{fig_ab} (c). The second variant is our region-guided selective fusion subnet without ER module, namely ``w/o ER", which removes the edge refinement module and generates the final prediction map by $4\times$ upsampling the prediction map $\mathbf{P}_f^1$, as shown in Fig~\ref{fig_ab} (d). The ablation study results are shown in Table.~\ref{tab:abl} and Table.~\ref{tab:ab2}.


From Table.~\ref{tab:abl}, it is obvious that our initial feature extraction subnet with only a single modality branch including ``Ours\_V", ``Ours\_D", and ``Ours\_T" outperforms the three baseline models including ``Base\_V", ``Base\_D", and ``Base\_T", which proves the strong feature extraction capability of our initial feature extraction subnet.  {Besides, from Table.~\ref{tab:ab_QRS}, we can find that our model performs better than ``w/o QA'', ``w/o HQ'', and ``w/o LQ'' in terms of all evaluation metrics, which demonstrates the effectiveness of the quality-aware region selection subnet. In addition, we can see that the performance of ``w/o HQ'' is lower than ``w/o LQ''. This indicates that high-quality regions can provide more contributions to improve the model. From the aforementioned descriptions, we can clearly demonstrate the effectiveness of each component in the quality-aware region selection subnet. Finally, according to Table.~\ref{tab:ab2}, it can be seen that our model performs better than ``w/o IIA'' and ``w/o ER'' in terms of all evaluation metrics, which proves the effectiveness of all components of region-guided selective fusion subnet. Generally speaking, from the above ablation studies, we can firmly prove the effectiveness and rationality of the design of our model.}


%
\begin{figure}[!t]
\centering
\includegraphics[width=0.48\textwidth]{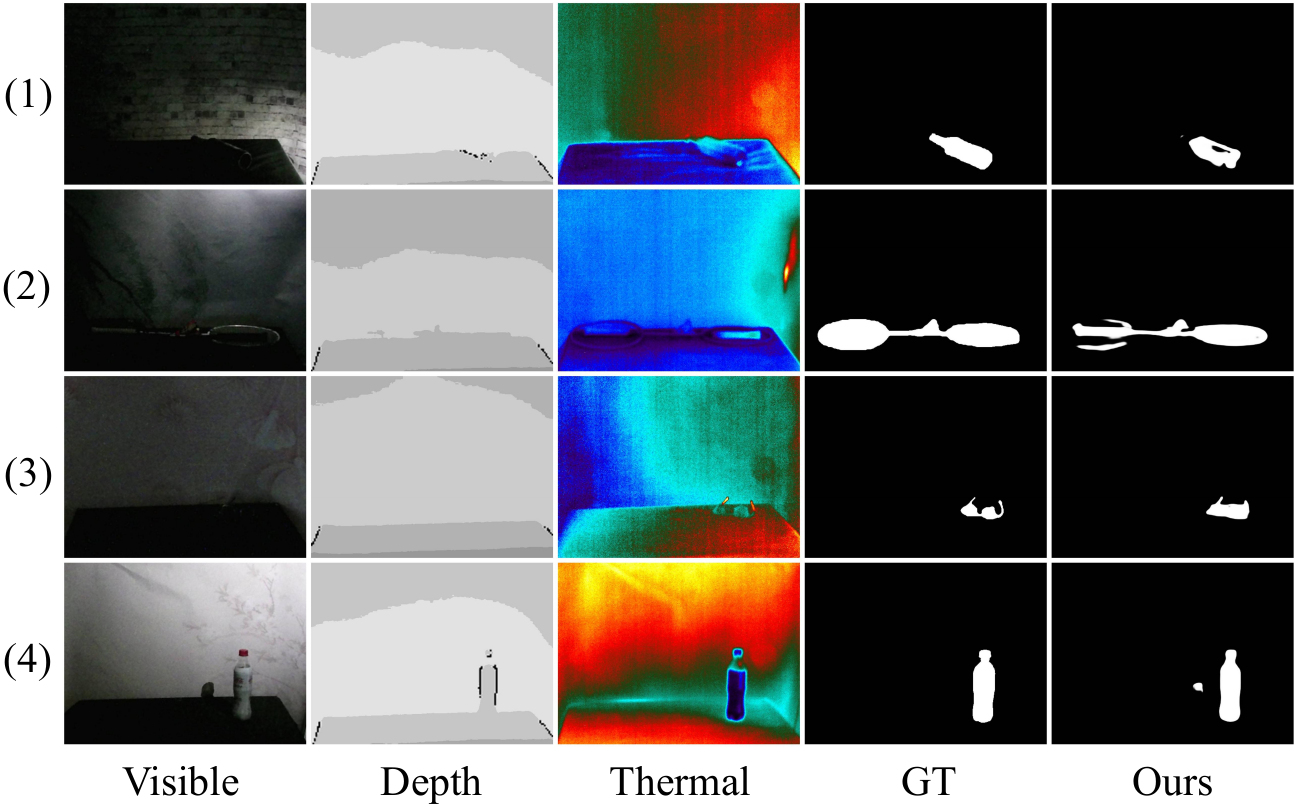}
\caption{ {Some failure examples.}} 
\label{fig_failure}
\end{figure}
 {
\subsection{Failure Cases and Analysis}
\label{sec:Failure}
As previously mentioned, the proposed QSFNet provides effective predictions results for V-D-T SOD task. However, our model still struggles to produce satisfactory results when dealing with some challenging scenes, as shown in Fig.~\ref{fig_failure}. For instance, in the $1^{st}$ and $2^{nd}$ rows of Fig.~\ref{fig_failure}, our model incorrectly omits a portion of the salient region, which shares a similar appearance and temperature with the background. Besides, in the $3^{rd}$ row of Fig.~\ref{fig_failure}, though our model accurately locates the body of salient object, the highlight of the boundary details within the object is rough and incorrect. Additionally, we find that the shadows of salient objects provide negative impact on the performance of our model. For the case in the $4^{th}$ row of Fig.~\ref{fig_failure}, our model mistakenly highlights the shaded part of the salient object. Besides, since our QSFNet is trained in three stages, it has a relatively high training cost, where we can see the large paramters in Table.~\ref{tab:cc}.
Therefore, in our feature work, we will attempt to design a more effective end-to-end network to perceive multi-modal qualities, promote the interaction among multi-modal features, and facilitate multi-modal fusion. In this way, we can achieve a balance between detection accuracy and computational cost.
}

\section{Conclusion}
\label{sec:conclusion}
In this paper, to give a sufficient and suitable aggregation of visible, depth, and thermal information, we propose a quality-aware selective fusion network (QSF-Net), 
which learned the favorable regions (\emph{i.e.}, quality-aware regions) and guided the selective fusion process through the selected regions. Firstly, the shrinkage pyramid structure based initial feature extraction subnet equipped with multi-scale fusion (MSF) module can not only generate initial multi-scale features but also give a coarse saliency prediction for each modality. Secondly, the weakly-supervised quality-aware region selection subnet first finds the high-quality and low-quality regions from the initial saliency predictions, and then generates the pseudo label, which can be used to train this subnet. Following this way, we can acquire the quality-aware maps. Finally, the region-guided selective fusion subnet utilizes the intra-modality and inter-modality attention (IIA) module to enhance the intra-modal feature and fuse the inter-modal feature under the guidance of the selected regions in the quality-aware map. After that, the edge refinement (ER) module is employed to further refine the boundary details. We conduct extensive experiments on VDT-2048 datasets, and the results demonstrate the effectiveness and superiority of the proposed QSF-Net when compared with the state-of-the-art models.  {However, our model also has some drawbacks, such as performance degradation in some challenging scenarios, large training costs, etc. Therefore, in our future work, we will attempt to construct a more effective end-to-end quality-aware multi-modality feature fusion network for the V-D-T salient object detection.}

\ifCLASSOPTIONcaptionsoff
  \newpage
\fi

\bibliographystyle{IEEEtran}
\bibliography{refs}
\end{document}